\documentclass{article}

    \usepackage[final]{neurips_2024}

\usepackage[utf8]{inputenc} 
\usepackage[T1]{fontenc}    
\usepackage{url}            
\usepackage{booktabs}       
\usepackage{amsfonts}       
\usepackage{nicefrac}       
\usepackage{microtype}      
\usepackage{xcolor}         
\usepackage{enumitem}
\usepackage{bm}
\usepackage{wrapfig}
\usepackage{adjustbox}
\usepackage{amsmath}
\usepackage{multicol}
\usepackage{multirow}
\usepackage{xcolor,colortbl}
\usepackage[colorlinks, citecolor=green]{hyperref}

\title{Inferring Neural Signed Distance Functions by Overfitting on Single Noisy Point Clouds through Finetuning Data-Driven based Priors}

\author{%
Chao Chen$^1$ \ \quad Yu-Shen Liu$^1$\thanks{The corresponding author is Yu-Shen Liu. This work was supported by National Key R\&D Program of China (2022YFC3800600), and the National Natural Science Foundation of China (62272263, 62072268), and in part by Tsinghua-Kuaishou Institute of Future Media Data.} \quad  \textbf{Zhizhong Han}$^2$\    \\
  $^1$School of Software, Tsinghua University, Beijing, China \\
  $^2$Department of Computer Science, Wayne State University, Detroit, USA \\
\texttt{chenchao19@tsinghua.org.cn}\quad \texttt{liuyushen@mails.tsinghua.edu.cn}\\
  \texttt{h312h@wayne.edu}}

\begin{document}

\maketitle

\begin{abstract}
It is important to estimate an accurate signed distance function (SDF) from a point cloud in many computer vision applications. The latest methods learn neural SDFs using either a data-driven based or an overfitting-based strategy. However, these two kinds of methods are with either poor generalization or slow convergence, which limits their capability under challenging scenarios like highly noisy point clouds. To resolve this issue, we propose a method to promote‌ pros of both data-driven based and overfitting-based methods for better generalization, faster inference, and higher accuracy in learning neural SDFs. We introduce a novel statistical reasoning algorithm in local regions which is able to finetune data-driven based priors without signed distance supervision, clean point cloud, or point normals. This helps our method start with a good initialization, and converge to a minimum in a much faster way. Our numerical and visual comparisons with the state-of-the-art methods show our superiority over these methods in surface reconstruction and point cloud denoising on widely used shape and scene benchmarks. The code is available at \url{https://github.com/chenchao15/LocalN2NM}.
\end{abstract}

\section{Introduction}
It is an important task to estimate an implicit function from a point cloud in computer graphics, computer vision, and robotics. An implicit function, such as a signed distance function (SDF), describes a continuous 3D distance field to indicate distances to the nearest surfaces at arbitrary locations. Since point clouds are easy to obtain, they are widely used as an information source to estimate SDFs, particularly without using normals that are not available for most scenarios. The challenge for SDF estimation mainly comes from the difficulty of bridging the gap between the discreteness of point clouds and the continuity of implicit functions.

Recent methods~\cite{mildenhall2020nerf,Oechsle2021ICCV,handrwr2020,zhizhongiccv2021matching,zhizhongiccv2021completing,takikawa2021nglod,DBLP:journals/corr/abs-2105-02788,rematasICML21} overcome this challenge using either a data-driven based or an overfitting-based strategy. To map a point cloud to a signed distance field, the data-driven based methods~\cite{Mi_2020_CVPR,Genova:2019:LST,jia2020learning,Liu2021MLS,tang2021sign,Peng2020ECCV,ErlerEtAl:Points2Surf:ECCV:2020,Lin_2021_CVPR,Williams_2019_CVPR,Tretschk2020PatchNets} rely on a prior learned with signed distance supervision from a large-scale dataset, while the overfitting-based methods~\cite{DBLP:conf/icml/GroppYHAL20,Atzmon_2020_CVPR,zhao2020signagnostic,atzmon2020sald,yifan2020isopoints,DBLP:journals/corr/abs-2106-10811,NeuralPoisson,Zhizhong2021icml,chibane2020neural,NEURIPS2023_c87bd584} do not need signed distance supervision and just use the point cloud to infer a signed distance field. However, both of the two kinds of methods have pros and cons. The data-driven based methods can do inference fast but suffers from the need of large-scale training samples and poor generalization to instances that are unseen during training. Although the overfitting-based methods have a better generalization ability and do not need the large-scale signed distance supervision, they usually require a much longer time to converge during inference. The cons of these two kinds of methods dramatically limit the performance of learning neural SDFs under challenging scenarios like highly noisy point clouds. Therefore, beyond pursuing higher accuracy of SDFs, how to balance the generalization ability and the convergence efficiency is also a significant issue.

To resolve this issue, we propose to learn an SDF from a single point cloud by finetuning data-driven based priors. Our key idea is to promote the advantages of both the data-driven based and the overfitting-based strategy to pursue better generalization, faster inference, and higher accuracy. Our method overfits a neural network on a single point cloud to estimate an SDF with a novel loss without using signed distance supervision, clean point, or point normals, where the neural network was pretrained as a data-driven based prior from large-scale signed distance supervision. With finetuning priors, our method can generalize better on unseen instances than the data-driven based methods, and also converge much more accurate SDFs in a much faster way than the overfitting-based methods. Moreover, our novel loss for finetuning the data-driven based prior can conduct a statistical reasoning in a local region which can recover more accurate and sharper underlying surface from noisy points. We report numerical and visual comparisons with the state-of-the-art methods and show our superiority over these methods in surface reconstruction and point cloud denoising on widely used shape and scene benchmarks. Our contributions are summarized below,

\begin{itemize}
\item We introduce a method which is capable of funetuning a data-driven based prior by minimizing an overfitting-based loss without signed distance supervision, leading to neural SDFs with better generalization, faster inference, and higher accuracy.
\item The proposed overfitting-based loss can conduct a novel statistical reasoning in local regions, which improves the accuracy of neural SDFs inferred from noisy point clouds.
\item Our method produces the state-of-the-art results in surface reconstruction and point cloud denoising on the widely used benchmarks.
\end{itemize}

\section{Related Works}
Learning implicit functions has achieved promising performance in various tasks~\cite{mildenhall2020nerf,Oechsle2021ICCV,handrwr2020,zhizhongiccv2021matching,zhizhongiccv2021completing,takikawa2021nglod,DBLP:journals/corr/abs-2105-02788,rematasICML21,Han2019ShapeCaptionerGCacmmm,DBLP:journals/corr/abs-2108-03743,Hu2023LNI-ADFP}. We can learn neural implicit representations from different supervision including 3D supervision~\cite{DBLP:journals/corr/abs-1901-06802,Park_2019_CVPR,MeschederNetworks,chen2018implicit_decoder}, multi-view images~\cite{sitzmann2019srns,DIST2019SDFRcvpr,Jiang2019SDFDiffDRcvpr,prior2019SDFRcvpr,shichenNIPS,DBLP:journals/cgf/WuS20,Volumetric2019SDFRcvpr,lin2020sdfsrn,yariv2020multiview,yariv2021volume,geoneusfu,neuslingjie,Yu2022MonoSDF,yiqunhfSDF,Vicini2022sdf,wang2022neuris}, and point clouds~\cite{Williams_2019_CVPR,liu2020meshing,Mi_2020_CVPR,Genova:2019:LST}. We briefly review the existing methods related to point clouds below.

\subsection{Data-driven based Methods}
In 3D supervision, many techniques utilize a data-driven approach to learning priors, and then apply these learned priors to infer implicit models for unseen point clouds. Some strategies focus on acquiring global priors~\cite{Mi_2020_CVPR,Genova:2019:LST,jia2020learning,Liu2021MLS,tang2021sign,Peng2020ECCV,ErlerEtAl:Points2Surf:ECCV:2020,Lin_2021_CVPR} at the shape level, whereas others aim to boost the generalization of these priors by learning local priors~\cite{Williams_2019_CVPR,Tretschk2020PatchNets,DBLP:conf/eccv/ChabraLISSLN20,jiang2020lig,Boulch_2022_CVPR,DBLP:conf/cvpr/MaLH22} at the component or patch level. These learned priors facilitate the marching cubes algorithm~\cite{Lorensen87marchingcubes} to reconstruct surfaces from implicit fields. The effectiveness of these methods often rely on extensive datasets, but they may not generalize well when facing with unseen point clouds that significantly deviate in geometry from training samples.

\subsection{Overfitting-based Methods}
In an effort to enhance generalization, some methods concentrate on precisely fitting neural networks to single point clouds. These methods incorporate innovative constraints~\cite{DBLP:conf/icml/GroppYHAL20,Atzmon_2020_CVPR,zhao2020signagnostic,atzmon2020sald,yifan2020isopoints,DBLP:journals/corr/abs-2106-10811,NeuralPoisson}, utilize gradients~\cite{Zhizhong2021icml,chibane2020neural,NEURIPS2023_c87bd584}, employ differentiable Poisson solvers~\cite{Peng2021SAP}, or apply specially tailored priors~\cite{DBLP:conf/cvpr/MaLH22,DBLP:conf/cvpr/MaLZH22} to learn either signed~\cite{Zhizhong2021icml,DBLP:conf/icml/GroppYHAL20,Atzmon_2020_CVPR,zhao2020signagnostic,atzmon2020sald,chaompi2022,BaoruiTowards,NeuralTPScvpr} or unsigned distance functions~\cite{chibane2020neural,Zhou2022CAP-UDF,Zhou_2023_ICCV}. Despite achieving significant advances, these approaches typically require clean point clouds to accurately determine distance or occupancy fields around the point clouds.

\subsection{Learning from Noisy Point Clouds}
The key to accurately reconstructing surfaces on noisy point clouds is to minimize the effect of noise in inferring implicit functions. PointCleanNet~\cite{DBLP:journals/cgf/RakotosaonaBGMO20} was developed to filter out noise from point clouds through a data-driven approach. GPDNet~\cite{DBLP:conf/eccv/PistilliFVM20} incorporated graph convolution based on dynamically generated neighborhood graphs to enhance noise reduction. Some other methods leveraged point cloud convolution~\cite{Boulch_2022_CVPR}, alternating latent topology~\cite{Wang_2023_CVPR,Mao_2024_CVPR}, semi-supervised strategy~\cite{Zhu_2024_WACV, chou2022gensdf}, dual and integrated latent~\cite{shim2024ditto}, or neural kernel field~\cite{Williams_2022_CVPR, Huang_2023_CVPR} to reduce noise from point clouds. On the unsupervised front, TotalDenoising~\cite{DBLP:conf/iccv/Casajus0R19} adopts principles similar to Noise2Noise~\cite{DBLP:conf/icml/LehtinenMHLKAA18}, utilizing a spatial prior suitable for unordered point clouds. DiGS~\cite{ben2021digs} employs a soft constraint for unoriented point clouds. Noise2NoiseMapping~\cite{BaoruiNoise2NoiseMapping} leverage statistical reasoning among multiple noisy point clouds with specially designed losses. Some methods using downsample-upsample frameworks~\cite{DBLP:conf/mm/LuoH20}, gradient fields~\cite{luo2021score,ShapeGF,Chen_2023_ICCV, ouasfi2024unsupervised,aminie2022}, convolution-free intrinsic occupancy network~\cite{ouasfi2023mixingdenoising}, intra-shape regularization~\cite{NEURIPS2023_525c95ff}, eikonal equation~\cite{NEURIPS2023_2d6336c1,fainstein2024dudf}, neural Galerkin~\cite{huang2022neuralgalerkin} and neural splines~\cite{DBLP:conf/cvpr/WilliamsTBZ21} have been implemented to further diminish noise in point clouds. Our method falls in this category, but we aim to promote the advantages of both the data-driven based and the overfitting-based strategy to pursue better generalization, faster inference, and higher accuracy.

\begin{figure}[t]
  \centering
   \vspace{-0.4cm}
  \includegraphics[width=0.99\linewidth]{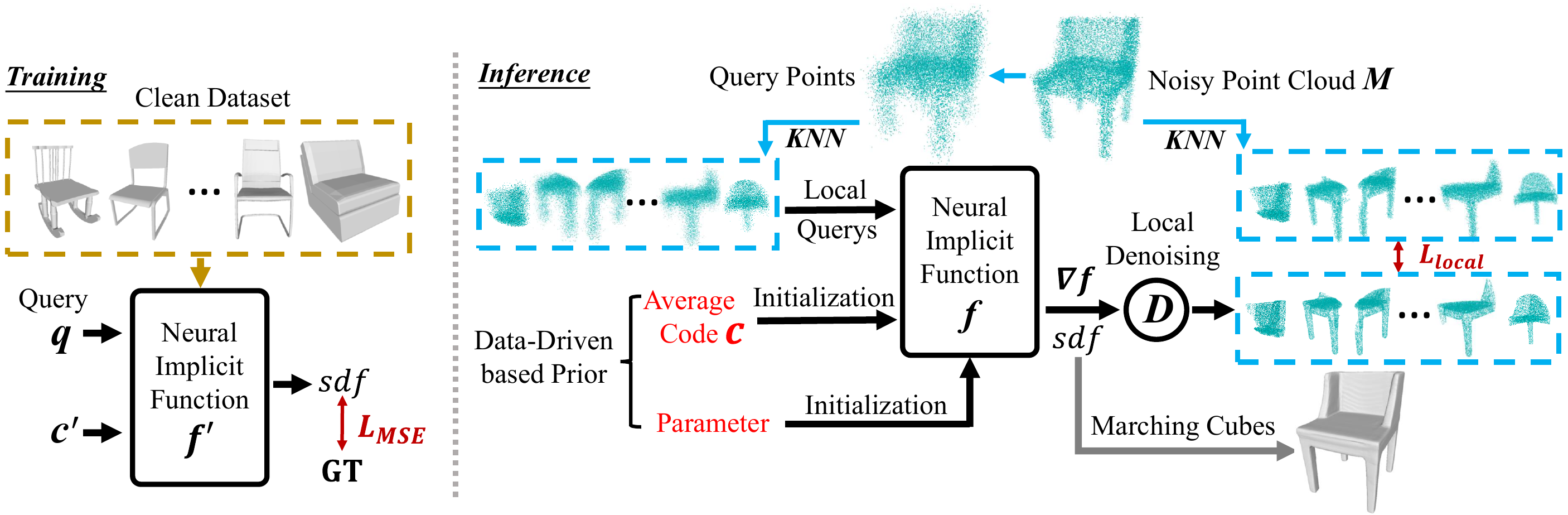}
  \vspace{-0.1in}
  \caption{The overview of our method. We learn the data-driven based prior by learning a neural implicit function $f'$ with a condition $\bm{c}'$ on a clean dataset. During inference, we employ a novel statistical reasoning algorithm to infer a neural SDF $f$ for a noisy point cloud $M$ with learned prior (average code and learned parameter). }
  \vspace{-0.6cm}
  \label{fig:overview}
\end{figure}

\section{Method}
\noindent\textbf{Overview. }We aim to infer a neural SDF $f$ from a single point cloud with noises $M$. Our method includes two stages shown in Fig.~\ref{fig:overview}, one is to learn a prior $f'$ in a data-driven manner, the other is to infer a neural SDF $f$ on unseen noisy point cloud $M$. At the first stage, we learn a prior by training a neural SDF using ground truth signed distances of clean meshes indicated by embeddings $\bm{c}_j'$. At the second stage, we finetune the learned prior $f'$ to infer a neural SDF $f$ of $M$ using our proposed local noise to noise mapping, where the embedding $\bm{c}$ indicating $M$ is also learned. We can use the marching cubes algorithm~\cite{Lorensen87marchingcubes} to extract the zero-level set of $f$ as the mesh surface of $M$.

\noindent\textbf{Neural Signed Distance Function. }We leverage an SDF $f$ to represent the geometry of a shape. An SDF $f$ is an implicit function that can predict a signed distance $s$ for an arbitrary location $q$, i.e., $s=f(q)$. The latest methods usually train a neural network to approximate an SDF from signed distance supervision or infer an SDF from 3D point clouds or multi-view images. A level set is an iso-surface formed by the points with the same signed distance value. For instance, zero-level set is a special level set, which is formed by points with a signed distance of 0. On the zero-level set, the gradient $\nabla f(q)$ of the SDF $f$ at an arbitrary location $q$ is also the surface normal at $q$.

\noindent\textbf{Data-driven Based Prior. }As shown in Fig.~\ref{fig:overview}, we employ an auto-decoder similar to DeepSDF~\cite{Park_2019_CVPR} for learning a prior $f'$ in a data-driven manner and inferring a neural SDF $f$ for single point clouds with noises, respectively. We employ a data-driven strategy to learn a prior $f’$ from clean meshes first. Specifically, we learn $f’$ with an embedding $\bm{c}_j'$ as a condition of queries. For each shape, we sample queries $q$ around a shape represented by $\bm{c}_j'$, and establish the signed distance supervision by recording the signed distance $s$ to the ground truth mesh. Thus, we learn the prior $f'$ by minimizing the prediction errors to the ground truth signed distances,

\begin{equation}
\label{Eq:error}
\min_{f',\{\bm{c}_i'\}} \sum_{i=1}^I\sum_{j=1}^J ||s_i^j-f’(q_j,\bm{c}_i')||_2^2+\alpha \sum_{i=1}^I ||\bm{c}_i'||_2^2,
\end{equation}

\noindent where $\bm{c}_i'$ is a learnable condition for the $i$-th training shape, $q_j$ is the $j$-th query that is randomly sampled around the $i$-th shape, and $s_i^j$ is the ground truth signed distance. We also add a regularization term on the learned embeddings $\bm{c}_i'$, and $\alpha$ is the balance weight.

\noindent\textbf{Signed Distance Inference. }With the learned prior $f’$, we infer a neural SDF $f$ for a single point cloud with noises $M$. We do not require ground truth signed distances, clean point clouds, or even point normal during the inference of $f$. Specifically, we infer $f$ by finetuning parameters of $f’$ with a learnable embedding $\bm{c}$ indicating the single point cloud with noises. The finetuning relies on a novel statistical reasoning algorithm on local regions. 

The advantage of our method lies in the capability of conducting the statistical reasoning in local regions. Comparing to the global reasoning method~\cite{BaoruiNoise2NoiseMapping}, our method is able to not only infer more accurate geometry but also significantly improve the efficiency. Our method starts from randomly sampling a local region $m_n$ on the shape $M$. We randomly select one point on $M$ as the center of $m_n$, and set up its $K$ nearest noisy points as a local region $m_n$. Then, we randomly sample $U$ queries $\{\bar{q}_u\}_{u=1}^U$ around $m_n$, and also randomly select $U$ noisy points $\{p_v\}_{v=1}^U$ out of $m_n$ for statistically reasoning the surface in each iteration. 

Our key idea of inferring a neural SDF $f$ is to estimate a mean zero-level set that is consistent to all points in the local region $m_n$. To this end, we use the $U$ sampled queries $\{\bar{q}_u\}$ to represent the zero-level set in this area using $f$, and minimize the distances of the $U$ noisy points $\{p_v\}$ to the zero-level set in each iteration. Statistically, the expectation of the zero-level set should have the minimum distance to all the noisy point splitting in region $m_n$. 

Specifically, we first project the $U$ sampled queries $\{\bar{q}_u\}$ onto the zero-level set of $f$ using a differentiable pulling operation~\cite{Zhizhong2021icml}. For each query $\bar{q}_u$, its projection on the zero-level set is,

\begin{equation}
\label{Eq:pulling}
\bar{q}_u'=\bar{q}_u-s*\nabla f(\bar{q}_u,\bm{c})/|\nabla f(\bar{q}_u,\bm{c})|,
\end{equation}

\noindent where $\bar{q}_u'$ is the projection of $\bar{q}_u$ on the zero-level set, $s=f(\bar{q}_u,\bm{c})$, $\nabla f(\bar{q}_u,\bm{c})$ is the gradient of $f$ at the location $\bar{q}_u$, and $\bm{c}$ is the learnable embedding that represents the noisy point cloud $M$. 

With the pulling operation, we can use projections $\{\bar{q}_u'\}$ of queries $\{\bar{q}_u\}$ to approximate the zero-level set in region $m_n$.
With a coarse zero-level set estimation, we expect this zero-level set can be consistent to various subsets of noises $\{p_v\}$ sampled from $m_n$. Thus, we minimize the errors between the $\{\bar{q}_u'\}_{u=1}^U$ and a subset of points $\{p_v\}_{v=1}^U$ on area $m_n$ in each optimization iteration,

\begin{equation}
\label{Eq:pulling}
\min_{f, \bm{c}} \mathbb{E}_{m_n\sim M, \bar{q}_u\sim m_n, p_v\sim m_n} EMD(\{\bar{q}_u'\},\{p_v\}) + \beta ||\bm{c}||_2^2,
\end{equation}

\noindent where we learn $f$ through finetuning the prior $f’$ and learning the embedding $\bm{c}$ representing the noisy point cloud $M$. The expectation is over the local regions $m_n$ that randomly sampled from the noisy point cloud $M$, and the subset patch $p_v$ randomly sampled from each $m_n$. We follow the method~\cite{BaoruiNoise2NoiseMapping} to use the EMD to evaluate the distance between the two sets of points, which leads the neural SDF $f$ to converge on the specific noisy point cloud $M$.

\noindent\textbf{Initialization. }The network architecture of $f$ is the same to the one of prior $f’$. We learn $f$ with the parameters of $f’$ as the initialization, representing the prior that we learned. For the embedding $\bm{c}$ that represents $M$, we initialize $\bm{c}$ as the center of the embedding space learned by the prior $f'$ in Eq.~\ref{Eq:error}, i.e., $\bm{c}=1/I\sum_{i=1}^I \bm{c}_i'$. This initialization is important for the accuracy and efficiency of learning $f$ for single noisy point cloud $M$. This finetuning of parameters of $f’$ also shows advantages over the auto-decoding~\cite{Park_2019_CVPR} in terms of generalization and efficiency. We will justify these advantages in our experiments.

\noindent\textbf{Implementation Details. }We randomly select one point from noisy point cloud $M$ as a center, and select its $K=1000$ nearest points to form a local region $m_n$. We also randomly sample $U=1000$ queries around the $K$ noisy points for statistically reasoning. Specifically, we adopt a method introduced by NeuralPull~\cite{Zhizhong2021icml} to sample queries around each one of the $K$ noisy points. We use a Gaussian distribution centered at each point and set the standard deviation as the distance to the 51th nearest neighbor in the point cloud. We run the marching cubes for surface reconstruction at a resolution of 256 for shapes, and 512 for large-scale scenes.

The length of the embedding $\bm{c}$ or $\bm{c}'$ is set to $256$. We use Adam optimizer for learning a neural implicit network, which is an auto-decoder similar to DeepSDF~\cite{Park_2019_CVPR}. For training, we use an initial embedding learning rate of 0.0005 for updating embeddings and an auto-decoder learning rate of $0.001$ for optimizing the prior network. Both learning rates are decreased by 0.5 for every 500 epochs. We train the prior network $f’$ for 2000 epochs. For inference, we finetune the network $f'$ for each noisy point cloud in $4000$ iterations with a learning rate of $0.0001$. 

\section{Experiments and Analysis}
We compare our method with the latest methods in terms of numerical and visual results on synthetic point clouds and real scans in surface reconstruction.

\textbf{Datasets and Metric. }We use eight datasets including shapes and scenes in the evaluations. For shapes, we conduct experiments under five datasets including ShapeNet~\cite{shapenet2015}, ABC~\cite{ErlerEtAl:Points2Surf:ECCV:2020}, FAMOUS~\cite{ErlerEtAl:Points2Surf:ECCV:2020}, Surface Reconstruction Benchmark (SRB)~\cite{Williams_2019_CVPR} and D-FAUST~\cite{dfaust:CVPR:2017}. For scenes, we conduct experiments under three real scan datasets including 3D Scene~\cite{DBLP:journals/tog/ZhouK13}, KITTI~\cite{Geiger2012CVPR}, Paris-rue-Madame~\cite{DBLP:conf/icpram/SernaMGD14}, and nuScenes~\cite{nuscenes}. We leverage L1 Chamfer Distance ($CD_{L1}$), L2 Chamfer Distance ($CD_{L2}$) to evaluate the error between the reconstructed surface and ground truth. We also use Normal Consistency (NC)~\cite{MeschederNetworks} and F-Score~\cite{Tatarchenko_2019_CVPR} with a threshold of 1\% to evaluate the normal accuracy of the reconstructed surface. In the ablation study, we also report time consumption to highlight the superiority of our data-driven based prior. For KITTI and Paris-rue-Madame datasets, due to their lack of ground truth meshes, we only report visual comparisons.

\begin{table}[tb]
\centering 
\resizebox{\textwidth}{!}{
    \begin{tabular}{c|c|c|c|c|c|c|c|c|c|c|c}
     \hline
     Metrics&PSR~\cite{DBLP:journals/tog/KazhdanH13}&PSG~\cite{DBLP:conf/cvpr/FanSG17}&R2N2~\cite{DBLP:conf/eccv/ChoyXGCS16}&COcc~\cite{DBLP:conf/eccv/PengNMP020}&SAP~\cite{Peng2021SAP}&OCNN~\cite{wang2020deep}&IMLS~\cite{Liu2021MLS}&POCO~\cite{pococvpr2022}&ALTO~\cite{Wang_2023_CVPR}&N2NM~\cite{BaoruiNoise2NoiseMapping}&Ours\\
     \hline
     $CD_{L1}$&0.299&0.147&0.173&0.044&0.034&0.067&0.031&0.030&0.028&0.026&\textbf{0.023}\\
     NC&0.772&-&0.715&0.938&0.944&0.932&0.944&0.950&0.955&0.962&\textbf{0.973}\\
     F-Score&0.612&0.259&0.400&0.942&0.975&0.800&0.983&0.984&0.985&0.991&\textbf{0.992}\\
     \hline
   \end{tabular}}
   \caption{Numerical Comparisons on ShapeNet dataset in terms of $CD_{L1}\times10$, NC and F-Score.}
   \vspace{-0.3in}
   \label{table:shapenet}
\end{table}

\begin{wrapfigure}{r}{0.7\linewidth}
\vspace{-0.2in}
\includegraphics[width=\linewidth]{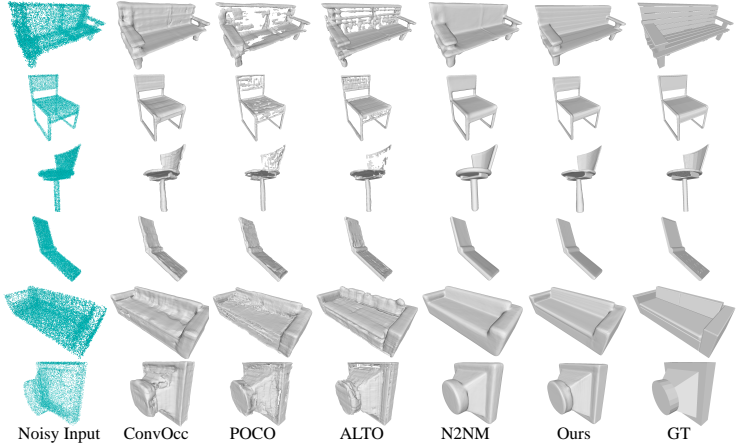}
\vspace{-0.25in}
\caption{\label{fig:shapenet}Comparison in surface reconstruction on ShapeNet. More visual results are provided in the appendix.}
\vspace{-0.2in}
\end{wrapfigure}

\subsection{Surface Reconstruction for Shapes} 

\textbf{Evaluation on ShapeNet. }We first report our results on shapes from ShapeNet. We report evaluations by comparing our method with the latest prior-based and overfitting-based methods in Tab~\ref{table:shapenet}. For prior-based methods, we compare our method with PSG~\cite{DBLP:conf/cvpr/FanSG17}, R2N2~\cite{DBLP:conf/eccv/ChoyXGCS16}, COcc~\cite{DBLP:conf/eccv/PengNMP020}, OCNN~\cite{wang2020deep}, IMLS~\cite{Liu2021MLS}, POCO~\cite{pococvpr2022}, and ALTO~\cite{Wang_2023_CVPR}. All of these methods are pretrained to learn priors using shapes with noises in training set of ShapeNet. We also follow these methods to use the same set of training shapes to learn our prior. For overfitting-based methods, we compare our method with PSR~\cite{DBLP:journals/tog/KazhdanH13}, SAP~\cite{Peng2021SAP}, and N2NM~\cite{BaoruiNoise2NoiseMapping}. These methods did not need to learn a prior, and have the ability of inferring neural implicit functions on each shape in the testing set. We also follow these methods and report our results by finetuning our prior through overfitting on each testing shape. All the shapes for testing are corrupted with noises with a variance of $0.005$. 

\begin{wraptable}{r}{0.32\linewidth}
\vspace{-0.05in}
\centering
\resizebox{\linewidth}{!}{
   \begin{tabular}{c|c|c|c}
     \hline
     Metrics&SAP~\cite{Peng2021SAP}&N2NM~\cite{BaoruiNoise2NoiseMapping}&Ours\\
     \hline
     Time&14 min&46 min&\textbf{5 min}\\
     \hline
   \end{tabular}}
   \vspace{-0.15in}
  \caption{Time consumption on ShapeNet dataset with overfitting-based methods.}
  \label{table:time}
  \vspace{-0.15in}
\end{wraptable}

The comparisons in Tab.~\ref{table:shapenet} indicate that our method can infer much more accurate neural implicit functions than the prior-based methods. The improvement comes from the ability of conducting test time optimization with the learned prior and inferring signed distances using the local noise to noise mapping. Moreover, our local statistical reasoning not only achieves better ability of recovering geometry from noisy points than overfitting-based methods but also significantly reduces the time complexity during the test time overfitting procedure with our prior. Different from prior-based methods, our ability of conducting test-time optimization with our local statistical reasoning loss can significantly improve the generalization ability on unseen shapes. Tab.~\ref{table:time} shows that our method can infer neural implicit functions on single shapes much faster than the overfitting-based methods. We also demonstrate our advantages in visual comparisons in Fig.~\ref{fig:shapenet}.


\begin{wrapfigure}{r}{0.6\linewidth}
\vspace{-0.45in}
\includegraphics[width=\linewidth]{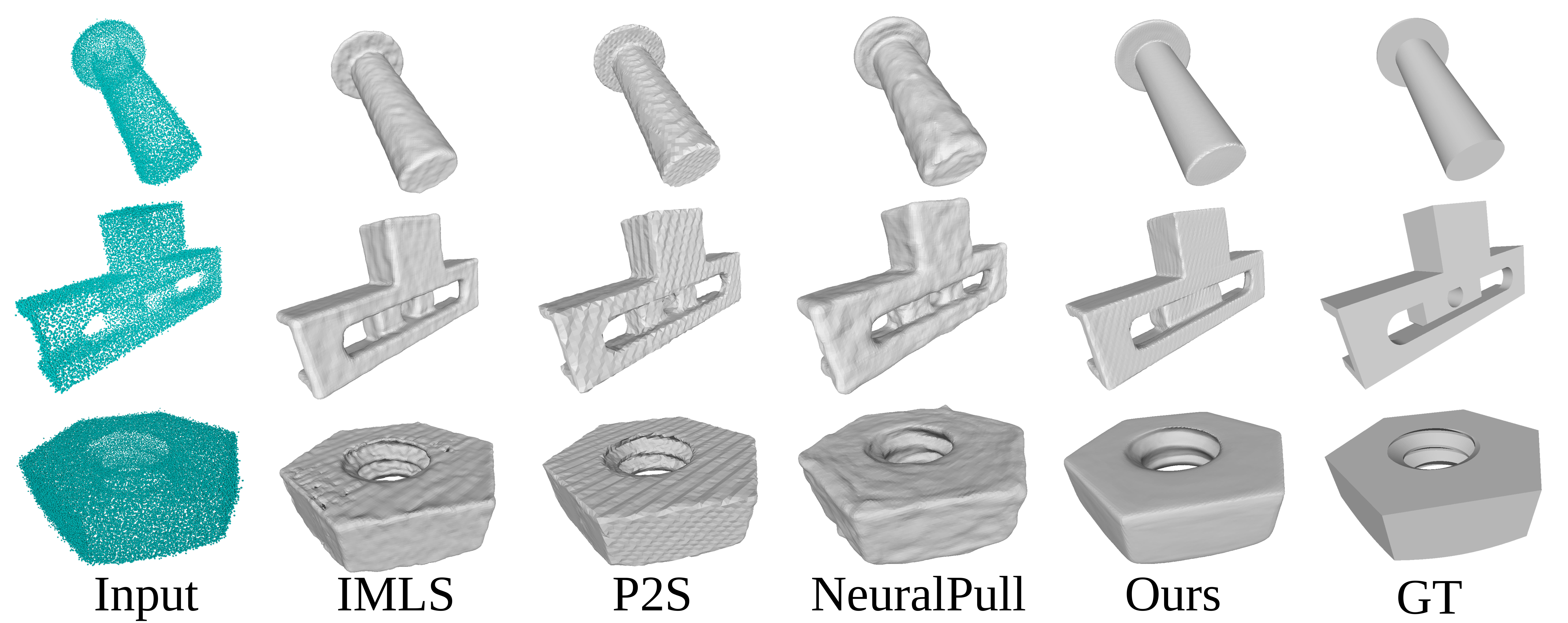}
\vspace{-0.3in}
\caption{\label{fig:abc}Comparison in surface reconstruction on ABC. More visual results are provided in the appendix.}
\vspace{-0.25in}
\end{wrapfigure}

\textbf{Evaluation on ABC. }
We also report our evaluations on ABC dataset in Tab.~\ref{table:abc}. We learn priors from shapes in training set, and finetune this prior for each single shape in the testing set. The numerical comparisons are conducted on the testing set of ABC dataset released by P2S~\cite{ErlerEtAl:Points2Surf:ECCV:2020}. It includes two versions with different noise levels. Similarly, we also report comparisons with prior-based methods and overfitting-based methods. With our local noise to noise mapping, we achieve the best performance over all baselines. Compared to prior-based methods, such as P2S~\cite{ErlerEtAl:Points2Surf:ECCV:2020}, COcc~\cite{DBLP:conf/eccv/PengNMP020}, and POCO~\cite{pococvpr2022}, our loss can infer more accurate geometry during the test time overfitting procedure. Also, the ability of finetuning the prior can also provide a coarse estimation and a good start for inferring neural implicit from single noisy points. Besides the accuracy, we also observe improvements on efficiency. Fig.~\ref{fig:abc} demonstrates the improvements over the baselines in terms of surface completeness and edge sharpness.

\begin{wrapfigure}{r}{0.55\linewidth}
\vspace{-0.25in}
\includegraphics[width=\linewidth]{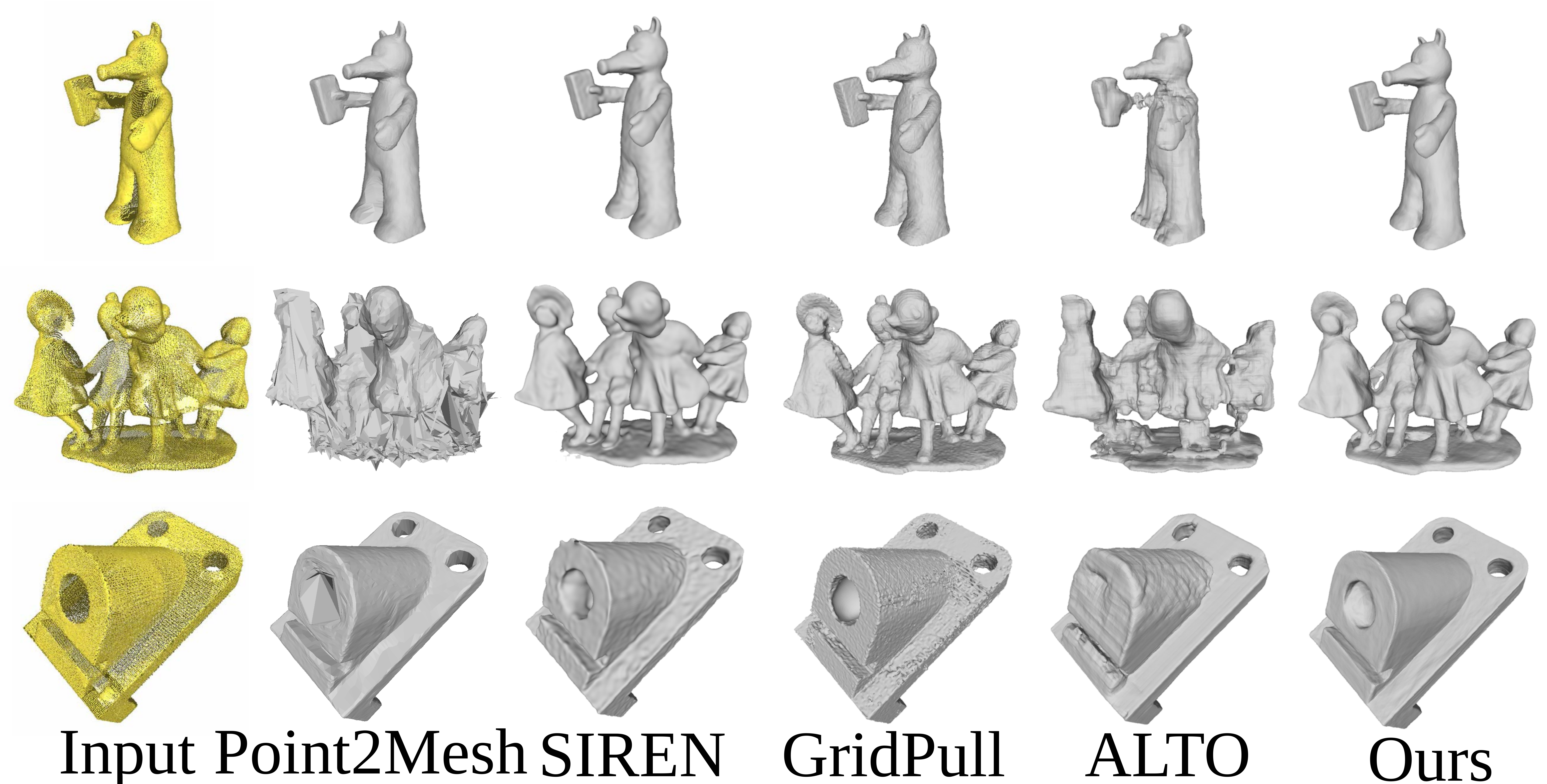}
\vspace{-0.25in}
\caption{\label{fig:srb}Comparison in surface reconstruction on SRB. More visual results are provided in the appendix.}
\vspace{-0.2in}
\end{wrapfigure}

\textbf{Evaluation on SRB. }We report previous experiments using man-made objects in ShapeNet and ABC dataset, We also report our results on real scans on SRB dataset~\cite{Williams_2019_CVPR}. Since there is no training samples on SRB, we use the prior learned from the ShapeNet as the prior for real scans. Although the shapes in ShapeNet are not similar to shapes in SRB, we found the prior can also work well with the scans on SRB. Different from the man-made objects, real scans have unknown noises. We report the evaluations with the prior-based and overfitting-based methods in Tab.~\ref{table:srb} and Fig.~\ref{fig:srb}. The comparisons show that our method achieves the best performance in implicit surface reconstruction. Under the same experimental settings, our method can infer more accurate geometry details with our local noise to noise mapping.


\begin{wrapfigure}{r}{0.6\linewidth}
\vspace{-0.35in}
\includegraphics[width=\linewidth]{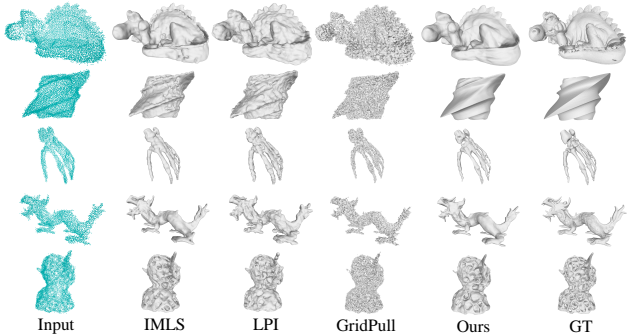}
\vspace{-0.25in}
\caption{\label{fig:famous}Comparison in surface reconstruction on FAMOUS. More visual results are provided in the appendix.}
\vspace{-0.23in}
\end{wrapfigure}

\textbf{Evaluation on FAMOUS. }We report evaluations on more complex shapes on FAMOUS dataset. Similar to SRB, we also use the prior learned from ShapeNet. We evaluate the performance on two kinds of noises in Tab.~\ref{table:famous}. We can see that our method can recover more geometry details and achieve higher accuracy and smoother surfaces. We also report visual comparisons in Fig.~\ref{fig:famous}, which also highlights our improvements in terms of accuracy, smoothness, completeness, and recovered sharp edges.

\begin{table}[tb]
\centering
\vspace{-0.1in}
\resizebox{\linewidth}{!}{
    \begin{tabular}{c|c|c|c|c|c|c|c|c|c|c}  
     \hline
          Dataset &PSR~\cite{DBLP:journals/tog/KazhdanH13} & P2S~\cite{ErlerEtAl:Points2Surf:ECCV:2020}&COcc~\cite{DBLP:conf/eccv/PengNMP020}&NP~\cite{Zhizhong2021icml}& IMLS~\cite{Liu2021MLS}&PCP~\cite{predictivecontextpriors2022}&POCO~\cite{pococvpr2022}&OnSurf~\cite{onsurfacepriors2022}&N2NM~\cite{BaoruiNoise2NoiseMapping}&Ours\\   %
     \hline
       ABC var& 3.29& 2.14&0.89&0.72 & 0.57&0.49&2.01&3.52&0.113&\textbf{0.096} \\ 
       ABC max& 3.89& 2.76&1.45&1.24& 0.68&0.57&2.50&4.30&0.139&\textbf{0.113} \\ 
     \hline
   \end{tabular}}
   \caption{Numerical Comparisons on ABC dataset in terms of $CD_{L2}\times100$.}
   \vspace{-0.1in}
   \label{table:abc}
\end{table}

\begin{table}[tb]
\centering
\vspace{-0.15in}
\resizebox{\linewidth}{!}{
    \begin{tabular}{c|c|c|c|c|c|c|c|c|c|c|c}
     \toprule
     Metrics&IGR~\cite{DBLP:conf/icml/GroppYHAL20}&Point2Mesh~\cite{DBLP:journals/tog/HanockaMGC20}&PSR~\cite{DBLP:journals/tog/KazhdanH13}&SIREN~\cite{sitzmann2019siren}&GP~\cite{Chen_2023_ICCV}&ALTO~\cite{Wang_2023_CVPR}&Steik~\cite{NEURIPS2023_2d6336c1}&SAP~\cite{Peng2021SAP}&NKSR~\cite{Huang_2023_CVPR}&N2NM~\cite{BaoruiNoise2NoiseMapping}&Ours\\
     \hline
     $CD_{L1}$&0.178&0.116&0.232&0.123&0.086&0.089&0.079&0.076&0.069&0.067&\textbf{0.055}\\
     F-Score&0.755&0.648&0.735&0.677&0.766&0.772&0.822&0.830&0.829&0.835&\textbf{0.860}\\
     \toprule
\end{tabular}
   }
   \caption{Numerical Comparisons on SRB dataset in terms of $CD_{L1}\times10$ and F-Score.}
   \vspace{-0.35in}
   \label{table:srb}
\end{table}

\begin{table}[t]
\centering
\vspace{-0.2in}
\resizebox{\linewidth}{!}{
    \begin{tabular}{c|c|c|c|c|c|c|c|c|c|c}  
     \hline
          Dataset &PSR~\cite{DBLP:journals/tog/KazhdanH13}&NP~\cite{Zhizhong2021icml}& IMLS~\cite{Liu2021MLS}&LPI~\cite{chaompi2022}&PCP~\cite{predictivecontextpriors2022}&POCO~\cite{pococvpr2022}&OnSurf~\cite{onsurfacepriors2022}&GP~\cite{Chen_2023_ICCV}&N2NM~\cite{BaoruiNoise2NoiseMapping}&Ours\\   %
     \hline
       F-var& 1.80& 0.28&0.80&0.19 & 0.07&1.50&0.59&0.13&0.033&\textbf{0.029} \\ 
       F-max& 3.41& 0.31&0.39&0.26& 0.30&2.75&3.64&0.21&0.117&\textbf{0.105} \\ 
     \hline
   \end{tabular}}
   \caption{Numerical Comparisons on FAMOUS dataset in terms of $CD_{L2}\times100$.}
   \label{table:famous}
   \vspace{-0.2in}
\end{table}

\begin{table}[tb]
\centering
\resizebox{0.8\linewidth}{!}{
    \begin{tabular}{c|c|c|c|c|c|c} 
     \toprule
     Metrics&IGR~\cite{DBLP:conf/icml/GroppYHAL20}&Point2Mesh~\cite{DBLP:journals/tog/HanockaMGC20}&PSR~\cite{DBLP:journals/tog/KazhdanH13}&SAP~\cite{Peng2021SAP}&N2NM~\cite{BaoruiNoise2NoiseMapping}&Ours\\
     \hline
     $CD_{L1}\times 10$&0.235&0.071&0.044&0.043&0.037&\textbf{0.034}\\
     F-Score&0.805&0.855&0.966&0.966&0.966&\textbf{0.973}\\
     NC&0.911&0.905&0.965&0.959&\textbf{0.970}&0.968\\
     \toprule
\end{tabular}
   }
   \caption{Accuracy of reconstruction on D-FAUST dataset in terms of $CD_{L1}$, NC and F-Score.}
   \vspace{-0.35in}
   \label{table:dfaust}
\end{table}

\begin{wrapfigure}{r}{0.6\linewidth}
\vspace{-0.67in}
\includegraphics[width=\linewidth]{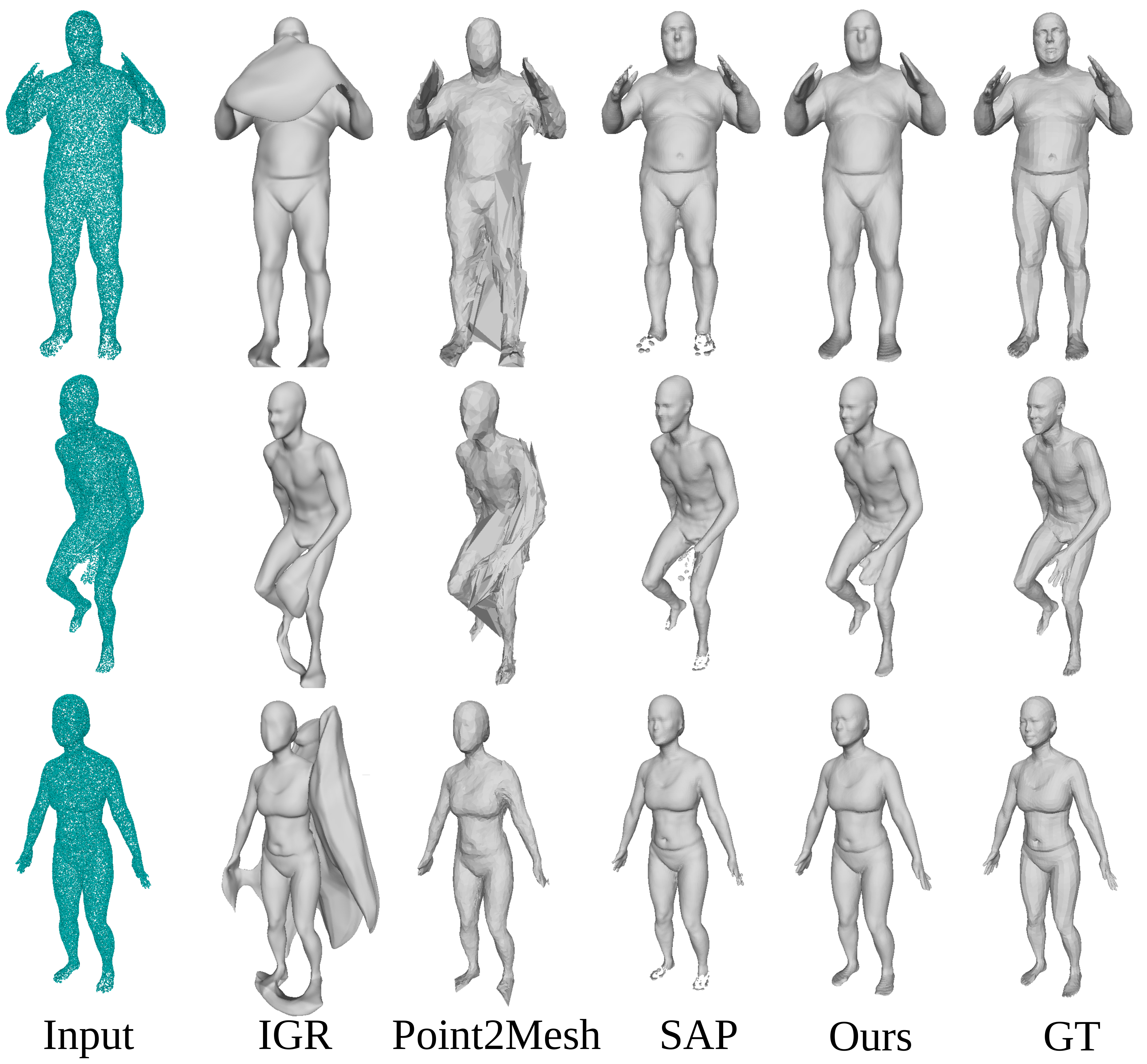}
\vspace{-0.25in}
\caption{\label{fig:dfaust}Comparison in surface reconstruction on D-FAUST. More visual results are provided in the appendix.}
\vspace{-0.5in}
\end{wrapfigure}

\textbf{Evaluation on D-FAUST. }Finally, we report our results on non-rigid shapes, i.e., humans. Different from rigid shapes in the previous experiments, humans are with more complex poses. We learn a prior from the training set, and finetuning the prior on unseen humans with different poses. We mainly compare our method with overfitting-based methods in Tab.~\ref{table:dfaust}. We can see that our method achieves the best performance in CD, F-Score, and comparable performance to N2NM~\cite{BaoruiNoise2NoiseMapping} but with faster inference speed. We further show the visual comparison in Fig.~\ref{fig:dfaust}. We can see that our method can recover more accurate geometry and poses.

\begin{table}[tb]
\vspace{-0.25in}
\centering
\resizebox{0.85\linewidth}{!}{
    \begin{tabular}{c|c|c|c|c|c|c}  
     \hline
     Metrics&COcc~\cite{DBLP:conf/eccv/PengNMP020}&LIG~\cite{jiang2020lig}&DeepLS\cite{DBLP:conf/eccv/ChabraLISSLN20}&NP~\cite{Zhizhong2021icml}&N2NM~\cite{BaoruiNoise2NoiseMapping}&Ours\\
     \hline
     $CD_{L2}\times 1000$&14.10&6.190&1.607&2.115&0.507&\textbf{0.389}\\
     $CD_{L1}$&0.052&0.048&0.025&0.034&0.019&\textbf{0.016}\\
     NC&0.908&0.849&0.915&0.900&0.929&\textbf{0.942}\\
     \hline
   \end{tabular}}
   \vspace{-0.0in}
   \caption{Numerical Comparisons on 3D Scene dataset in terms of $CD_{L1}$, $CD_{L2}$ and NC. Detailed comparisons for each scene are provided in the appendix.}
   \label{table:3DScene}
   \vspace{-0.1in}
\end{table}

\begin{figure}[tb]
\vspace{-0.2in}
  \centering
   \includegraphics[width=\linewidth]{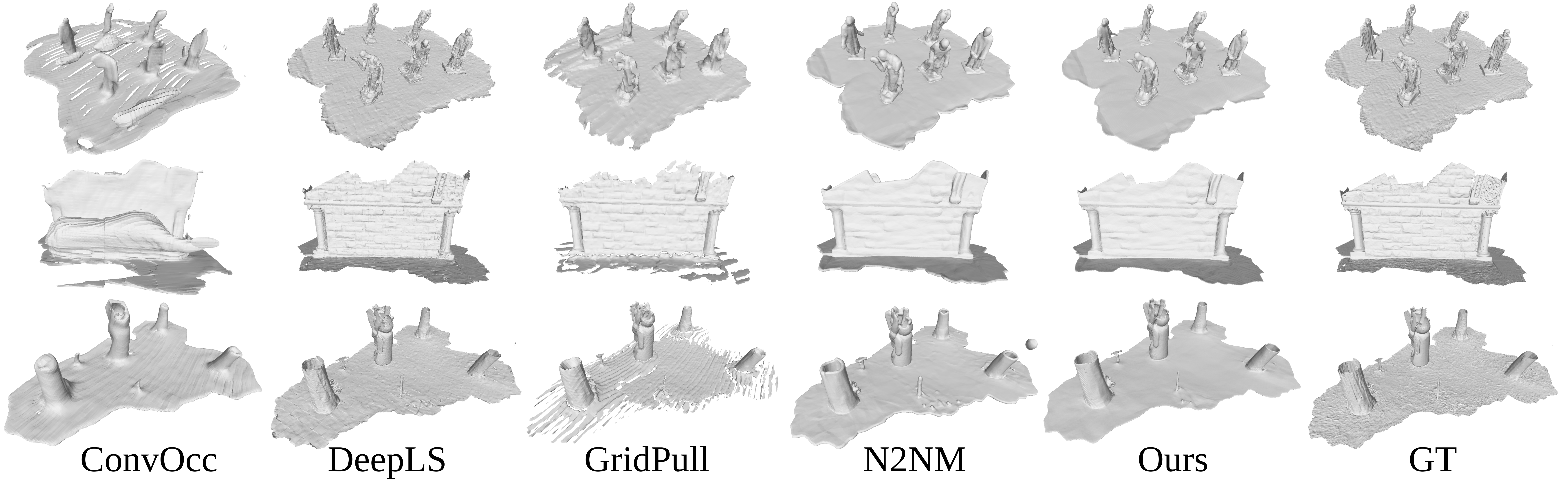}
\vspace{-0.25in}
\caption{\label{fig:3dsceneshort}Comparison in surface reconstruction on 3D Scene.}
\vspace{-0.1in}
\end{figure}

\subsection{Surface Reconstruction for Scenes}
Since we have a limited number of scenes for training, we use the prior learned from ShapeNet as the pretrained prior in our experiments on scenes. Specifically, we conduct experiments on four different scene datasets: 3D Scene~\cite{DBLP:journals/tog/ZhouK13}, KITTI~\cite{Geiger2012CVPR}, Paris-rue-Madame~\cite{DBLP:conf/icpram/SernaMGD14} and nuScenes~\cite{nuscenes}, where the results on nuScenes are reported in the appendix.

\textbf{Evaluation on 3D Scene. }We further evaluate our method in surface reconstruction for scenes in 3D Scene~\cite{DBLP:journals/tog/ZhouK13}. We follow previous methods LIG~\cite{jiang2020lig} to randomly sample $1000$ points per $m^2$. We compare our method with the latest methods including COcc~\cite{DBLP:conf/eccv/PengNMP020} and LIG~\cite{jiang2020lig}, DeepLS~\cite{DBLP:conf/eccv/ChabraLISSLN20}, NeuralPull (NP)~\cite{Zhizhong2021icml} and Noise2NoiseMapping (N2NM)~\cite{BaoruiNoise2NoiseMapping}. For prior-based methods COcc~\cite{DBLP:conf/eccv/PengNMP020} and LIG~\cite{jiang2020lig}, we leverage their released pretrained models to produce the results, and we also provide them with the ground truth point normals. For overfitting-based methods DeepLS~\cite{DBLP:conf/eccv/ChabraLISSLN20}, NP~\cite{Zhizhong2021icml} and N2NM~\cite{BaoruiNoise2NoiseMapping}, we overfit them to produce results with the same noisy point clouds. We follow LIG~\cite{jiang2020lig} to report $CD_{L1}$, $CD_{L2}$ and NC for evaluation. We report the comparisons in Tab.~\ref{table:3DScene}. The results demonstrate that our method outperforms both kinds of methods with learned priors such as LIG~\cite{jiang2020lig} and overfitting-based N2NM~\cite{BaoruiNoise2NoiseMapping}. The visual comparisons in Fig.~\ref{fig:3dsceneshort} show that our method can reveal more geometry details on real scans, which justifies our capability of handling noise in point clouds.

\begin{figure}[tb]
\vspace{-0.18in}
  \centering
   \includegraphics[width=\linewidth]{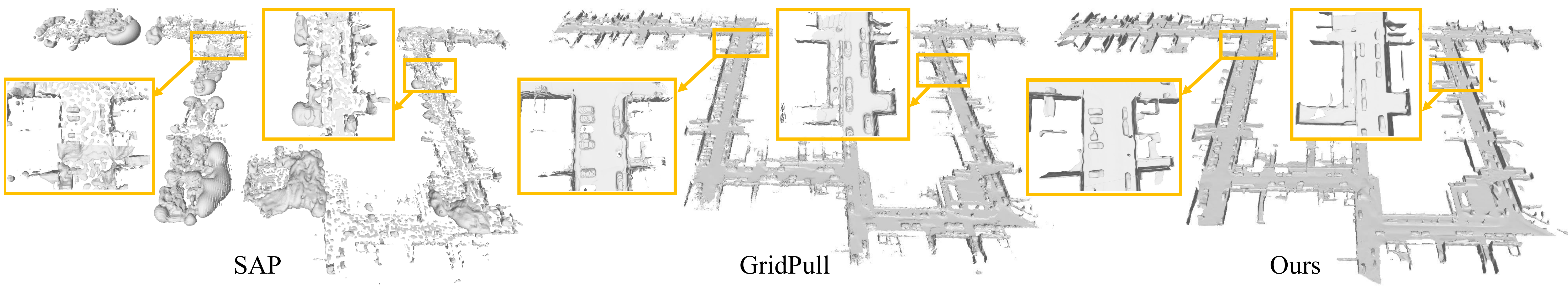}
\vspace{-0.27in}
\caption{\label{fig:kitti}Comparison in surface reconstruction on KITTI.}
\vspace{-0.1in}
\end{figure}

\textbf{Evaluation on KITTI. }Following GridPull~\cite{Chen_2023_ICCV}, we further evaluate our method on KITTI~\cite{Geiger2012CVPR} odometry dataset (Sequence 00, frame 3000 to 4000), which contains about 13.8 million points, which are split into 15 chunks. We reconstruct each of them and concatenate them together for visualization. We compare our method with the latest methods SAP~\cite{Peng2021SAP} and GridPull~\cite{Chen_2023_ICCV}. As shown in Fig.~\ref{fig:kitti}, our method is robust to noise in real scans, successfully generalizes to large-scale scenes, and achieves visual-appealing reconstructions with more details.

\begin{figure}[t]
\vspace{-0.1in}
  \centering
   \includegraphics[width=\linewidth]{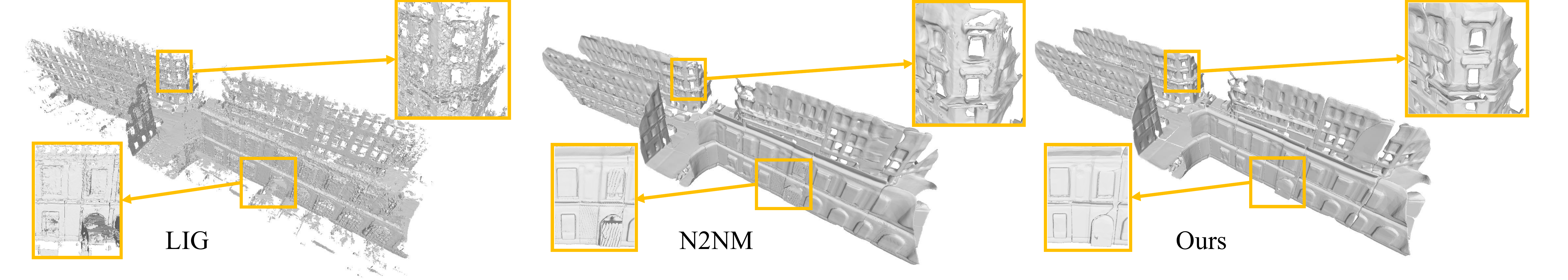}
\vspace{-0.23in}
\caption{\label{fig:Paris}Comparison in surface reconstruction on Paris-rue-Madame.}
\vspace{-0.32in}
\end{figure}

\textbf{Evaluation on Paris-rue-Madame. }Following N2NM~\cite{BaoruiNoise2NoiseMapping}, we further evaluate our method on Paris-rue-Madame~\cite{DBLP:conf/icpram/SernaMGD14}, which contains much noises. We split the $10$ million points into 50 chunks each of which is used to learn a neural implicit function. We compare our method with LIG~\cite{jiang2020lig} and N2NM~\cite{BaoruiNoise2NoiseMapping}. For LIG~\cite{jiang2020lig}, we produce the results for each chunk with released pretrained models. For N2NM~\cite{BaoruiNoise2NoiseMapping}, we overfit on all chunks until convergence. As shown in Fig.~\ref{fig:Paris}, we achieve better performance over LIG~\cite{jiang2020lig} and N2NM~\cite{BaoruiNoise2NoiseMapping} in large-scale surface reconstruction, which highlight our advantages in reconstructing complete and detailed surfaces from noisy scene point clouds.

\subsection{Ablation Studies}
We conduct ablation studies on the ABC dataset~\cite{ErlerEtAl:Points2Surf:ECCV:2020} to justify each module of our method.

\begin{wraptable}{r}{0.37\linewidth}
\vspace{-0.15in}
\centering
\resizebox{\linewidth}{!}{
   \begin{tabular}{c|c|c|c}  
    \hline
    Metric&128&256&512\\
    \hline
    $CD_{L2}\times100$&0.102&\textbf{0.096}&0.114\\
    \hline
  \end{tabular}}
   \vspace{-0.1in}
  \caption{Effect of the embedding size.}  
  \label{table:es}
  \vspace{-0.2in}
\end{wraptable}

\textbf{Embedding Size. }We evaluate our performance on different sizes of embedding $\bm{c}$. We try several sizes $\{128, 256, 512\}$ to infer the signed distance functions from a noisy point cloud. The numerical comparison in Tab.~\ref{table:es} shows that the optimal result is obtained with a size of $256$. Deviations from this value, either longer or shorter dimensions, leads to worse results with the current number of training samples.

\begin{wraptable}{r}{0.6\linewidth}
\vspace{-0.15in}
\centering
\resizebox{\linewidth}{!}{
   \begin{tabular}{c|c|c|c|c}
    \hline
    Metric&Without Prior&Without Embed&Fixed Param&With Prior\\
    \hline
    $CD_{L2}\times100$&0.108&0.103&0.144&\textbf{0.096}\\
    Time&1h&12min&30min&\textbf{8 min}\\
    \hline
  \end{tabular}}
   \vspace{-0.1in}
  \caption{Effect of the prior.}
  \label{table:prior}
  \vspace{-0.2in}
\end{wraptable}

\textbf{Prior. }We conduct experiments to explore the importance of data-driven based prior. We first replace our learned embedding $\bm{c}$ and parameter with randomly initialized embedding and parameter, or only replace $\bm{c}$ with randomly initialized embedding. As shown in Tab.~\ref{table:prior}, The degenerated result of ``Without Prior'' and ``Without Embed'' indicates that directly inferring implicit functions without our prior or learned embedding makes it difficult to accurately learn the surfaces of the noisy point clouds, and also slows the convergence. Then we fix the learned parameters and only optimize the embedding $\bm{c}$, similar to auto-decoding. The results also get worse, as shown in ``Fixed Param''.

\begin{wraptable}{r}{0.56\linewidth}
\vspace{-0.15in}
\centering
\resizebox{\linewidth}{!}{
   \begin{tabular}{c|c|c|c}
    \hline
    Metric&Voxel&Sphere (Fixed Size)&Sphere (KNN)\\
    \hline
    $CD_{L2}\times100$&0.314&0.101&\textbf{0.096}\\
    \hline
  \end{tabular}}
   \vspace{-0.1in}
  \caption{Effect of splitting strategies.}
  \label{table:split}
  \vspace{-0.2in}
\end{wraptable}

\textbf{Local Region Splitting. }We further validate the effectiveness of local region splitting strategies. We employ three different splitting strategies in Tab.~\ref{table:split}. We first split the whole space where the noisy point cloud is located uniformly into multiple voxel blocks, as shown by the result of ``Voxel''. The severely degenerated results indicate that this splitting strategy is even worse than the global method N2NM~\cite{BaoruiNoise2NoiseMapping}, as it results in many empty voxel blocks. Then we randomly select a point from the noisy point cloud as a center to sample all points within a radius of $0.1$ as a local region. The result of ``Sphere (Fixed Size)'' slightly degenerates due to some of the spheres containing too few points. In contrast, our splitting strategy, as shown by the result of ``Sphere (KNN)'', ensures that each local region has enough points to help achieve superior performance.

\begin{wraptable}{r}{0.34\linewidth}
\vspace{-0.0in}
\centering
\resizebox{\linewidth}{!}{
   \begin{tabular}{c|c|c}
    \hline
    Metric&Global&Local\\
    \hline
    $CD_{L2}\times100$&0.106&\textbf{0.096}\\
    Time&21 min&\textbf{8 min}\\
    \hline
  \end{tabular}}
   \vspace{-0.1in}
  \caption{Effect of local mapping.}
  \label{table:local}
  \vspace{-0.15in}
\end{wraptable}

\textbf{Global and Local. }With our learned prior, we compare our performance in global and local mappings with finetuning the priors. We report results obtained with the local noise to noise mapping or the global one during the finetuning. As shown in Tab.~\ref{table:local}, the numerical comparison shows that the global mapping struggles to infer local details from noisy point clouds. Moreover, our local prior also converges faster than the global statistical reasoning.

\begin{wraptable}{r}{0.47\linewidth}
\vspace{-0.15in}
\centering
\resizebox{\linewidth}{!}{
   \begin{tabular}{c|c|c|c|c}
    \hline
    Metric&500&1000&3000&5000\\
    \hline
    $CD_{L2}\times100$&0.102&\textbf{0.096}&0.111&0.114\\
    \hline
  \end{tabular}}
   \vspace{-0.1in}
  \caption{Effect of local region size.}
  \label{table:regionsize}
  \vspace{-0.2in}
\end{wraptable}

\textbf{Local Region Size. }We further validate the effectiveness of local region sizes (points number in a local region) in Tab.~\ref{table:regionsize}. We use different local region sizes including $\{500, 1000, 3000, 5000\}$. The results show that $1000$ is the best. 


\begin{wraptable}{r}{0.55\linewidth}
\vspace{-0.15in}
\centering
\resizebox{\linewidth}{!}{
    \begin{tabular}{c|c|c|c|c}  
     \hline
          Metric & Random & Square & Sphere (SAL) & Ours \\   %
     \hline
       Time&8.3min&7.1min&5.5min&\textbf{5.0min}\\ 
     \hline
   \end{tabular}}
   \caption{The effect of SDF initialization.}
   \label{table:sdfinit}
   \vspace{-0.05in}
\end{wraptable}

\begin{wrapfigure}{r}{0.6\linewidth}
  \centering
  \vspace{-0.35in}
   \includegraphics[width=\linewidth]{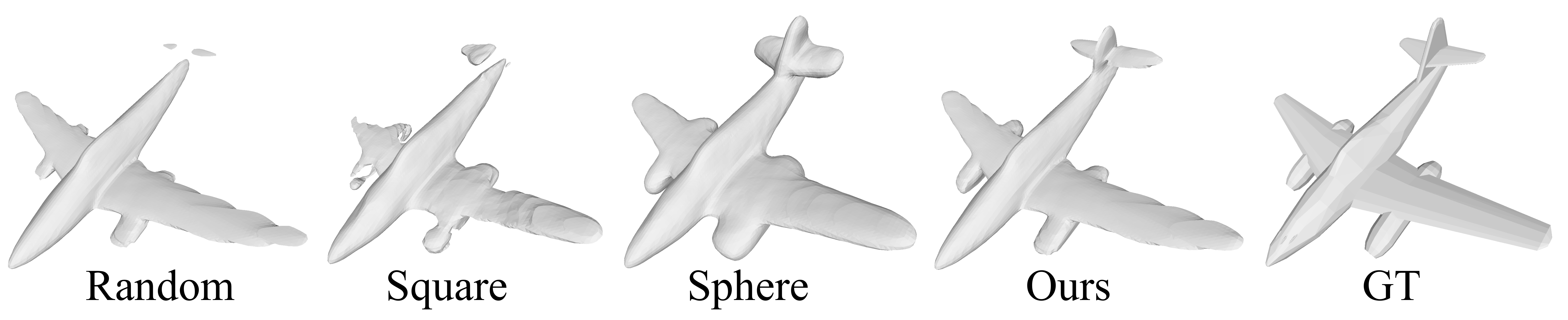}
   \vspace{-0.25in}
\caption{\label{fig:sdfinit}Comparison with different SDF initializations.
}
\vspace{-0.25in}
\end{wrapfigure}

\textbf{SDF initialization. }We further validate the effectiveness of different SDF initializations in Tab.~\ref{table:sdfinit} and Fig.~\ref{fig:sdfinit}, including random initialization, geometry initialization~\cite{Atzmon_2020_CVPR}, initialization to a simple square shape, and ours. 

We can see our prior can reconstruct more accurate surfaces from single noisy point clouds in much shorter time than any other initializations.

\begin{figure}[h]
  \centering
  \vspace{-0.1in}
   \includegraphics[width=0.7\linewidth]{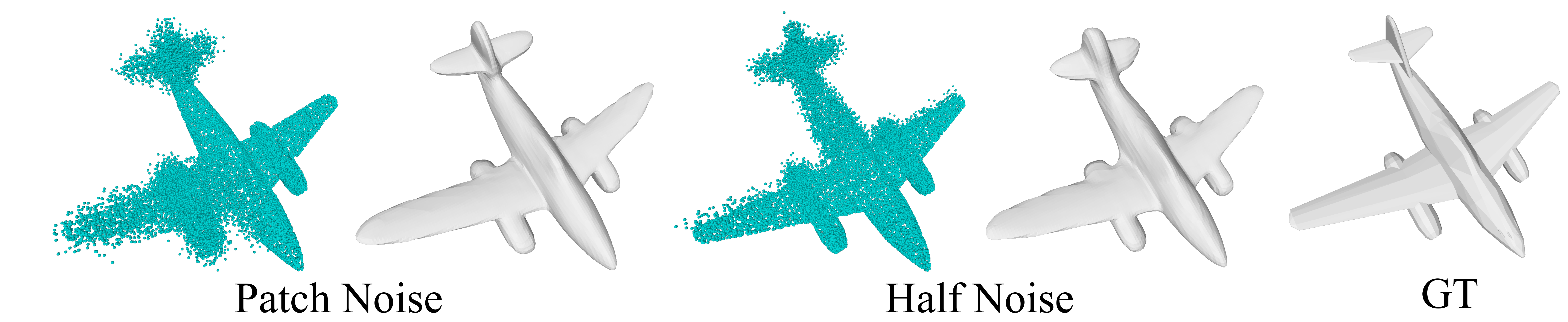}
   \vspace{-0.1in}
\caption{\label{fig:nonuniformnoise}Visual results with nonuniform noises.
}
\vspace{-0.1in}
\end{figure}

\begin{wrapfigure}{r}{0.55\linewidth}
  \centering
  \vspace{-0.15in}
   \includegraphics[width=\linewidth]{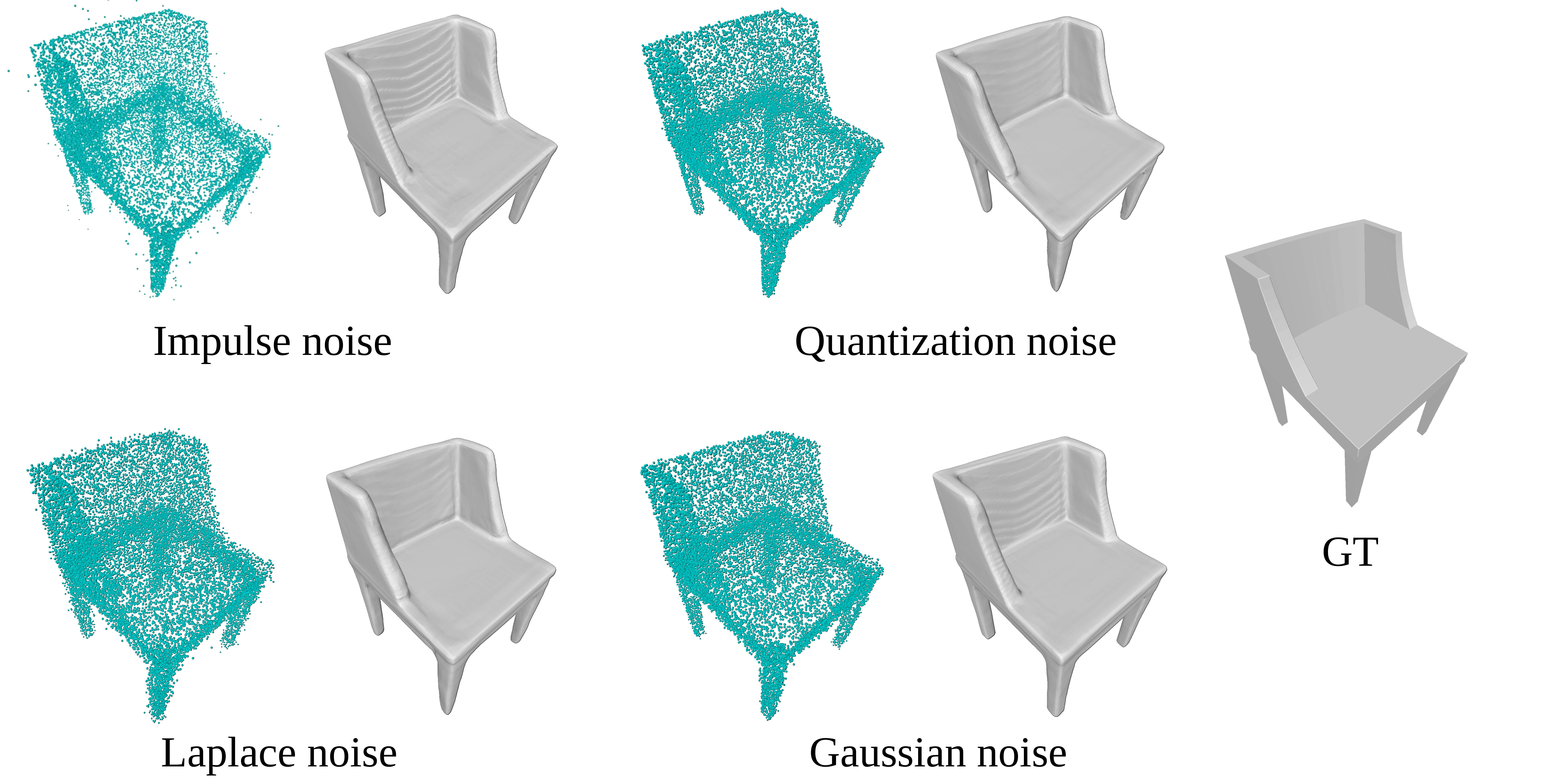}
   \vspace{-0.2in}
\caption{\label{fig:noisetype}Visual results with different noise types.
}
\vspace{-0.2in}
\end{wrapfigure}

\textbf{Noise Type. }We report our performance with various noise types, i.e., impulse noise, quantization noise, Laplacian noise, and Gaussian noise. Visual comparison in Fig.~\ref{fig:noisetype} justifies that we can also handle other types of noise quite well. Moreover, we also tried more challenging cases with nonuniform noises which do not have a zero expectation across a shape, like a shape with only a half of points having noises or a shape with several patches having noises. The result in Fig.~\ref{fig:nonuniformnoise} shows that our method can also handle nonuniform noises well.


\begin{wrapfigure}{r}{0.7\linewidth}
  \centering
  \vspace{-0.05in}
   \includegraphics[width=\linewidth]{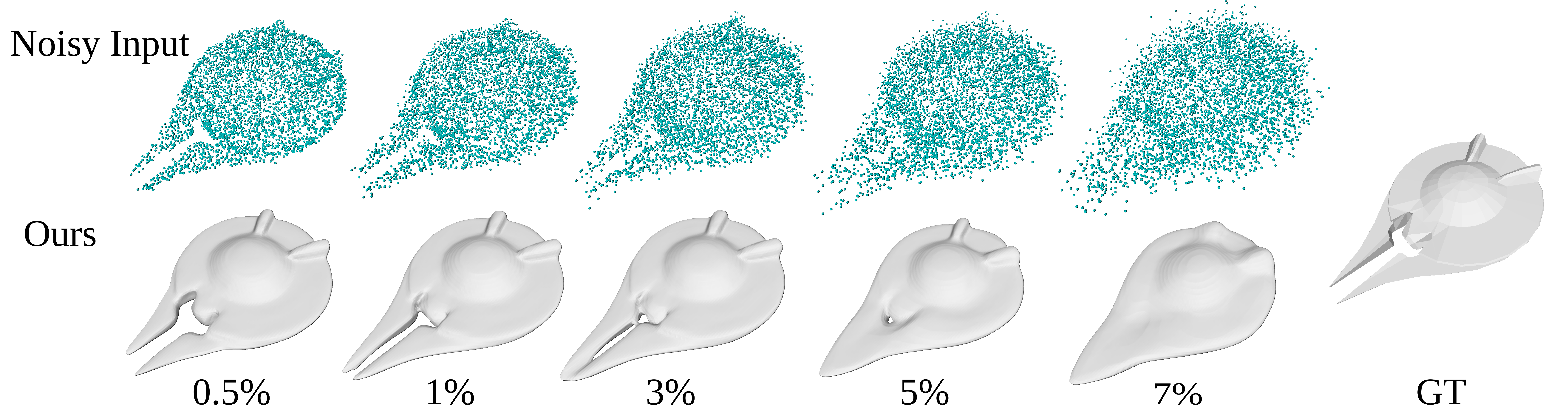}
   \vspace{-0.3in}
\caption{\label{fig:noiselevel}Visual comparison with different noise levels.
}
\vspace{-0.2in}
\end{wrapfigure}

\textbf{Noise Level. }We report our performance on point clouds with different levels of noise. As shown in Tab.~\ref{table:noiselevel}, the noise levels of middle and max come from the ABC dataset~\cite{ErlerEtAl:Points2Surf:ECCV:2020}. The middle indicates noises with a variance of $0.01L$, where $L$ is the longest edge of the bounding box. The max indicates noises with a variance of $0.05L$. Our extreme noise comes with a variance of $0.07L$. 

\begin{wraptable}{r}{0.37\linewidth}
\vspace{-0.15in}
\centering
\resizebox{\linewidth}{!}{
   \begin{tabular}{c|c|c|c}
    \hline
    Method&Middle&Max&Extreme\\
    \hline
    N2NM~\cite{BaoruiNoise2NoiseMapping}&0.113&0.139&0.156\\
    Ours&\textbf{0.096}&\textbf{0.113}&\textbf{0.125}\\
    \hline
  \end{tabular}}
   \vspace{-0.1in}
  \caption{Effect of noise level.}
  \label{table:noiselevel}
  \vspace{-0.2in}
\end{wraptable}

The $CD_{L2}$ comparison shows that our results slightly degenerate with max and extreme noise, but still outperform N2NM~\cite{BaoruiNoise2NoiseMapping}. The visual results in Fig.~\ref{fig:noiselevel} indicates that our method is more robust to noises even when the noise variance is as large as 7\%.

\begin{wraptable}{r}{0.35\linewidth}
\vspace{-0.15in}
\centering
\resizebox{\linewidth}{!}{
    \begin{tabular}{c|c|c|c}  
     \hline
          Method & 25\% & 50\% & 100\%\\   %
     \hline
       N2NM~\cite{BaoruiNoise2NoiseMapping}&0.154&0.133&0.113\\ 
       Ours&\textbf{0.121}&\textbf{0.107}&\textbf{0.096} \\ 
     \hline
   \end{tabular}}
   \caption{Effect of sparsity.}
   \label{table:sparsity}
   \vspace{-0.1in}
\end{wraptable}  

\begin{figure}[h]
  \centering
  \vspace{-0.15in}
   \includegraphics[width=0.8\linewidth]{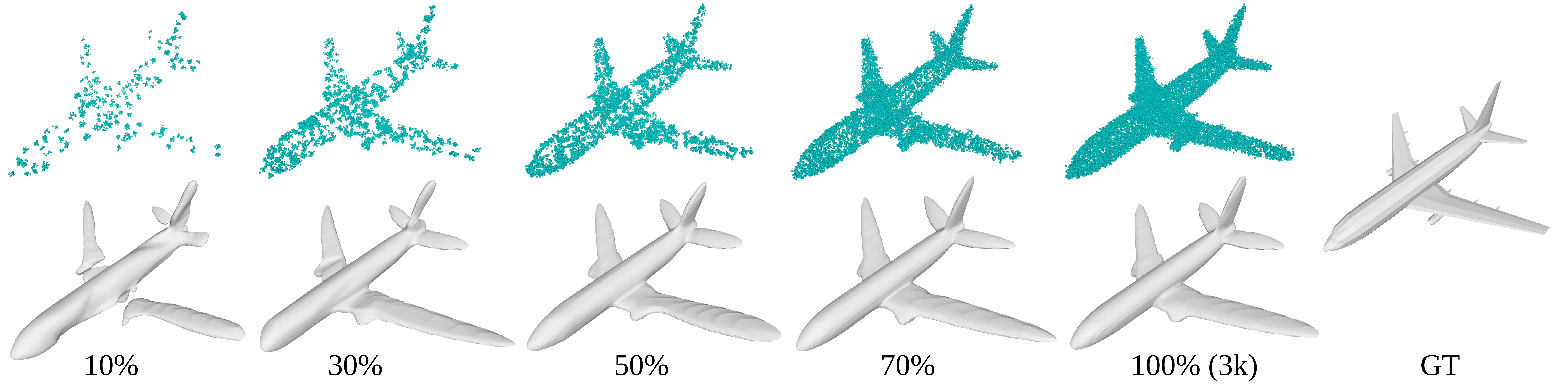}
   \vspace{-0.15in}
\caption{\label{fig:time}Visual comparison with different point numbers.
}
 \vspace{-0.15in}
\end{figure}

\textbf{Sparsity. }We report the effect of the sparsity of noisy point clouds. We downsample the noisy point clouds to 25\% and 50\% of their original size to validate the impact of sparsity. The $CD_{L2}$ results in Tab.~\ref{table:sparsity} and visual comparisons in Fig.~\ref{fig:time} indicate that our method can handle sparsity in noisy point clouds better than N2NM~\cite{BaoruiNoise2NoiseMapping}. Since our data-driven based prior can help to learn a more complete surface and reduce the impacts brought by the sparsity.

\begin{wraptable}{r}{0.55\linewidth}
\centering
\vspace{-0.15in}
\resizebox{\linewidth}{!}{
    \begin{tabular}{c|c|c|c|c|c}  
     \hline
          Metric & 10\% & 30\% & 50\% & 70\% & 100\% (3k)\\   %
     \hline
       Time&3.1min&3.6min&4.0min&4.5min&5.0min\\ 
     \hline
   \end{tabular}}
   \caption{The comparison of time consumption with different point numbers.}
   \label{table:time1}
    \vspace{-0.2in}
\end{wraptable}  

\textbf{Time Consumption. }Since our method can handle sparsity and require less time as the point number decreases, we conduct an experiment with downsampled noisy points in Tab.~\ref{table:time1}. Fig.~\ref{fig:time} indicates that we can work well on much fewer points, and also provide an alternative of improving efficiency.

\begin{wrapfigure}{r}{0.8\linewidth}
\vspace{-0.2in}
  \centering
   \includegraphics[width=\linewidth]{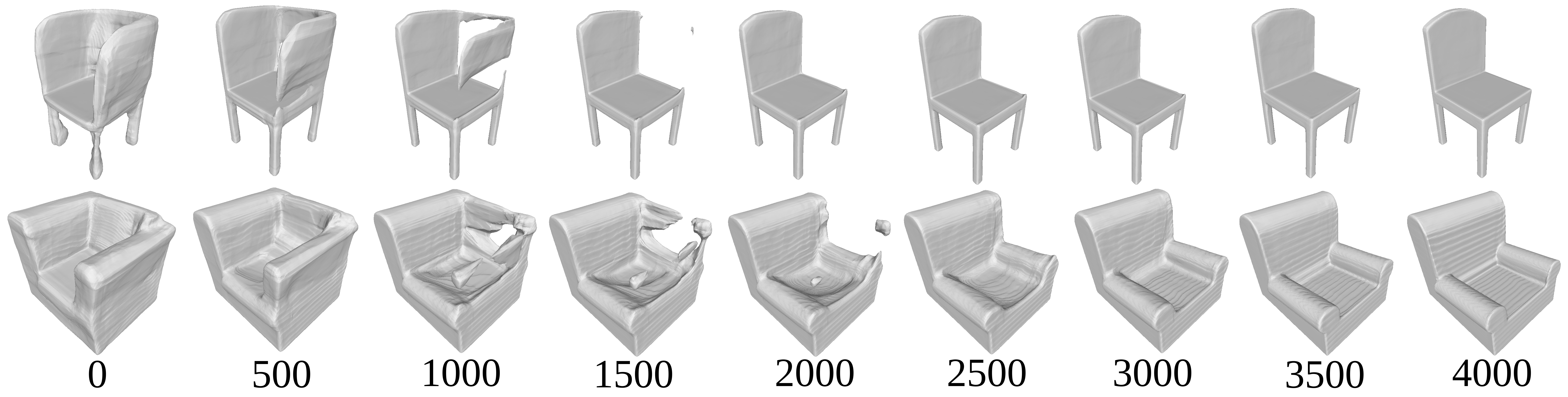}
  \vspace{-0.30in}
\caption{\label{fig:famous_app}Optimization during inference.}
\vspace{-0.1in}
\end{wrapfigure}

\textbf{Optimization. }We visualize the optimization process in Fig.~\ref{fig:famous_app}. We reconstruct meshes using the neural SDF $f$ learned in different iterations. We see that the shape is updated progressively to the ground truth shapes.

\section{Conclusion}
We propose a method to resolve the key problem in inferring SDFs from a single noisy point cloud. Our method can effectively use a data-driven based prior as an initialization, and infer a neural SDF by overfitting on a single noisy point cloud. The novel statistical reasoning successfully infers an accurate and smooth signed distance field around the single noisy point cloud with the data-driven based prior. By finetuning data-driven based priors with statistical reasoning, our method significantly improves the robustness, the scalability, the efficiency, and the accuracy in inferring SDFs from single point clouds. Our experimental results and ablations studies show our superiority and justify the effectiveness of the proposed modules.

{\small
\bibliographystyle{plain}
\bibliography{papers}

\begin{thebibliography}{100}

\bibitem{Atzmon_2020_CVPR}
Matan Atzmon and Yaron Lipman.
\newblock {SAL}: Sign agnostic learning of shapes from raw data.
\newblock In {\em IEEE Conference on Computer Vision and Pattern Recognition}, 2020.

\bibitem{atzmon2020sald}
Matan Atzmon and yaron Lipman.
\newblock {SALD:} sign agnostic learning with derivatives.
\newblock In {\em International Conference on Learning Representations}, 2021.

\bibitem{ben2021digs}
Yizhak Ben-Shabat, Chamin Hewa~Koneputugodage, and Stephen Gould.
\newblock {DiGS}: Divergence guided shape implicit neural representation for unoriented point clouds.
\newblock In {\em {IEEE} Conference on Computer Vision and Pattern Recognition}, 2022.

\bibitem{DBLP:journals/corr/abs-2106-10811}
Yizhak Ben{-}Shabat, Chamin~Hewa Koneputugodage, and Stephen Gould.
\newblock {DiGS} : Divergence guided shape implicit neural representation for unoriented point clouds.
\newblock {\em CoRR}, abs/2106.10811, 2021.

\bibitem{dfaust:CVPR:2017}
Federica Bogo, Javier Romero, Gerard Pons-Moll, and Michael~J. Black.
\newblock Dynamic {FAUST}: {R}egistering human bodies in motion.
\newblock In {\em IEEE Computer Vision and Pattern Recognition}, 2017.

\bibitem{Boulch_2022_CVPR}
Alexandre Boulch and Renaud Marlet.
\newblock {POCO}: Point convolution for surface reconstruction.
\newblock In {\em Proceedings of the IEEE/CVF Conference on Computer Vision and Pattern Recognition (CVPR)}, pages 6302--6314, June 2022.

\bibitem{pococvpr2022}
Alexandre Boulch and Renaud Marlet.
\newblock Poco: Point convolution for surface reconstruction.
\newblock In {\em {IEEE} Conference on Computer Vision and Pattern Recognition}, 2022.

\bibitem{nuscenes}
Holger Caesar, Varun Bankiti, Alex~H. Lang, Sourabh Vora, Venice~Erin Liong, Qiang Xu, Anush Krishnan, Yu~Pan, Giancarlo Baldan, and Oscar Beijbom.
\newblock nuscenes: A multimodal dataset for autonomous driving.
\newblock In {\em Proceedings of the IEEE/CVF Conference on Computer Vision and Pattern Recognition}, 2020.

\bibitem{ShapeGF}
Ruojin Cai, Guandao Yang, Hadar Averbuch-Elor, Zekun Hao, Serge Belongie, Noah Snavely, and Bharath Hariharan.
\newblock Learning gradient fields for shape generation.
\newblock In {\em European Conference on Computer Vision}, 2020.

\bibitem{DBLP:conf/iccv/Casajus0R19}
Pedro~Hermosilla Casajus, Tobias Ritschel, and Timo Ropinski.
\newblock Total denoising: Unsupervised learning of 3d point cloud cleaning.
\newblock In {\em {IEEE} International Conference on Computer Vision}, pages 52--60, 2019.

\bibitem{DBLP:conf/eccv/ChabraLISSLN20}
Rohan Chabra, Jan~Eric Lenssen, Eddy Ilg, Tanner Schmidt, Julian Straub, Steven Lovegrove, and Richard~A. Newcombe.
\newblock Deep local shapes: Learning local {SDF} priors for detailed {3D} reconstruction.
\newblock In {\em European Conference on Computer Vision}, volume 12374, pages 608--625, 2020.

\bibitem{shapenet2015}
Angel~X. Chang, Thomas Funkhouser, Leonidas Guibas, Pat Hanrahan, Qixing Huang, Zimo Li, Silvio Savarese, Manolis Savva, Shuran Song, Hao Su, Jianxiong Xiao, Li~Yi, and Fisher Yu.
\newblock {Shape{N}et: An Information-Rich 3D Model Repository}.
\newblock Technical Report arXiv:1512.03012 [cs.GR], Stanford University --- Princeton University --- Toyota Technological Institute at Chicago, 2015.

\bibitem{NeuralTPScvpr}
Chao Chen, Zhizhong Han, and Yu-Shen Liu.
\newblock Unsupervised inference of signed distance functions from single sparse point clouds without learning priors.
\newblock In {\em Proceedings of the IEEE/CVF Conference on Computer Vision and Pattern Recognition}, pages 17712--17723, 2023.

\bibitem{zhizhongiccv2021matching}
Chao Chen, Zhizhong Han, Yu-Shen Liu, and Matthias Zwicker.
\newblock Unsupervised learning of fine structure generation for {3D} point clouds by {2D} projections matching.
\newblock In {\em {IEEE} International Conference on Computer Vision}, 2021.

\bibitem{chaompi2022}
Chao Chen, Yu-Shen Liu, and Zhizhong Han.
\newblock Latent partition implicit with surface codes for 3d representation.
\newblock In {\em European Conference on Computer Vision}, 2022.

\bibitem{Chen_2023_ICCV}
Chao Chen, Yu-Shen Liu, and Zhizhong Han.
\newblock {GridPull}: Towards scalability in learning implicit representations from 3d point clouds.
\newblock In {\em Proceedings of the IEEE/CVF International Conference on Computer Vision (ICCV)}, pages 18322--18334, 2023.

\bibitem{chen2018implicit_decoder}
Zhiqin Chen and Hao Zhang.
\newblock Learning implicit fields for generative shape modeling.
\newblock {\em IEEE Conference on Computer Vision and Pattern Recognition}, 2019.

\bibitem{chibane2020neural}
Julian Chibane, Aymen Mir, and Gerard Pons-Moll.
\newblock Neural unsigned distance fields for implicit function learning.
\newblock {\em arXiv}, 2010.13938, 2020.

\bibitem{chou2022gensdf}
Gene Chou, Ilya Chugunov, and Felix Heide.
\newblock {GenSDF}: Two-stage learning of generalizable signed distance functions.
\newblock In {\em Advances in Neural Information Processing Systems}, pages 24905--24919, 2022.

\bibitem{DBLP:conf/eccv/ChoyXGCS16}
Christopher~B. Choy, Danfei Xu, JunYoung Gwak, Kevin Chen, and Silvio Savarese.
\newblock {3D}-r2n2: {A} unified approach for single and multi-view 3d object reconstruction.
\newblock In Bastian Leibe, Jiri Matas, Nicu Sebe, and Max Welling, editors, {\em European Conference on Computer Vision}, volume 9912, pages 628--644, 2016.

\bibitem{NeuralPoisson}
Angela Dai and Matthias Nie{\ss}ner.
\newblock Neural {Poisson}: Indicator functions for neural fields.
\newblock {\em arXiv preprint arXiv:2211.14249}, 2022.

\bibitem{ErlerEtAl:Points2Surf:ECCV:2020}
Philipp Erler, Paul Guerrero, Stefan Ohrhallinger, Niloy~J. Mitra, and Michael Wimmer.
\newblock {Points2Surf}: Learning implicit surfaces from point clouds.
\newblock In {\em European Conference on Computer Vision}, 2020.

\bibitem{fainstein2024dudf}
Miguel Fainstein, Viviana Siless, and Emmanuel Iarussi.
\newblock {DUDF}: Differentiable unsigned distance fields with hyperbolic scaling.
\newblock {\em arXiv preprint arXiv:2402.08876}, 2024.

\bibitem{DBLP:conf/cvpr/FanSG17}
Haoqiang Fan, Hao Su, and Leonidas~J. Guibas.
\newblock A point set generation network for {3D} object reconstruction from a single image.
\newblock In {\em 2017 {IEEE} Conference on Computer Vision and Pattern Recognition}, pages 2463--2471, 2017.

\bibitem{geoneusfu}
Qiancheng Fu, Qingshan Xu, Yew{-}Soon Ong, and Wenbing Tao.
\newblock {Geo-Neus}: Geometry-consistent neural implicit surfaces learning for multi-view reconstruction.
\newblock 2022.

\bibitem{Geiger2012CVPR}
Andreas Geiger, Philip Lenz, and Raquel Urtasun.
\newblock Are we ready for autonomous driving? the kitti vision benchmark suite.
\newblock In {\em Computer Vision and Pattern Recognition}, 2012.

\bibitem{Genova:2019:LST}
Kyle Genova, Forrester Cole, Daniel Vlasic, Aaron Sarna, William~T. Freeman, and Thomas Funkhouser.
\newblock Learning shape templates with structured implicit functions.
\newblock In {\em International Conference on Computer Vision}, 2019.

\bibitem{DBLP:conf/icml/GroppYHAL20}
Amos Gropp, Lior Yariv, Niv Haim, Matan Atzmon, and Yaron Lipman.
\newblock Implicit geometric regularization for learning shapes.
\newblock In {\em International Conference on Machine Learning}, volume 119 of {\em Proceedings of Machine Learning Research}, pages 3789--3799, 2020.

\bibitem{handrwr2020}
Zhizhong Han, Chao Chen, Yu-Shen Liu, and Matthias Zwicker.
\newblock {DRWR}: A differentiable renderer without rendering for unsupervised 3{D} structure learning from silhouette images.
\newblock In {\em International Conference on Machine Learning}, 2020.

\bibitem{Han2019ShapeCaptionerGCacmmm}
Zhizhong Han, Chao Chen, Yu-Shen Liu, and Matthias Zwicker.
\newblock {ShapeCaptioner}: Generative caption network for {3D} shapes by learning a mapping from parts detected in multiple views to sentences.
\newblock In {\em ACM International Conference on Multimedia}, 2020.

\bibitem{DBLP:journals/corr/abs-2108-03743}
Zhizhong Han, Xiyang Wang, Yu{-}Shen Liu, and Matthias Zwicker.
\newblock Hierarchical view predictor: Unsupervised 3d global feature learning through hierarchical prediction among unordered views.
\newblock In {\em Proceedings of the 29th ACM International Conference on Multimedia}, pages 3862--–3871, 2021.

\bibitem{DBLP:journals/tog/HanockaMGC20}
Rana Hanocka, Gal Metzer, Raja Giryes, and Daniel Cohen{-}Or.
\newblock Point2mesh: a self-prior for deformable meshes.
\newblock {\em {ACM} Transactions on Graphics}, 39(4):126, 2020.

\bibitem{Hu2023LNI-ADFP}
Pengchong Hu and Zhizhong Han.
\newblock Learning neural implicit through volume rendering with attentive depth fusion priors.
\newblock In {\em Advances in Neural Information Processing Systems}, 2023.

\bibitem{huang2022neuralgalerkin}
Jiahui Huang, Hao-Xiang Chen, and Shi-Min Hu.
\newblock A neural galerkin solver for accurate surface reconstruction.
\newblock {\em ACM Trans. Graph.}, 41(6), 2022.

\bibitem{Huang_2023_CVPR}
Jiahui Huang, Zan Gojcic, Matan Atzmon, Or~Litany, Sanja Fidler, and Francis Williams.
\newblock Neural kernel surface reconstruction.
\newblock In {\em Proceedings of the IEEE/CVF Conference on Computer Vision and Pattern Recognition}, pages 4369--4379, 2023.

\bibitem{jia2020learning}
Meng Jia and Matthew Kyan.
\newblock Learning occupancy function from point clouds for surface reconstruction.
\newblock {\em arXiv}, 2010.11378, 2020.

\bibitem{jiang2020lig}
Chiyu Jiang, Avneesh Sud, Ameesh Makadia, Jingwei Huang, Matthias Nie{\ss}ner, and Thomas Funkhouser.
\newblock Local implicit grid representations for {3D} scenes.
\newblock In {\em IEEE Conference on Computer Vision and Pattern Recognition}, 2020.

\bibitem{Jiang2019SDFDiffDRcvpr}
Yue Jiang, Dantong Ji, Zhizhong Han, and Matthias Zwicker.
\newblock {SDFDiff}: Differentiable rendering of signed distance fields for {3D} shape optimization.
\newblock In {\em IEEE Conference on Computer Vision and Pattern Recognition}, 2020.

\bibitem{DBLP:journals/tog/KazhdanH13}
Michael~M. Kazhdan and Hugues Hoppe.
\newblock Screened poisson surface reconstruction.
\newblock {\em {ACM} Transactions on Graphics}, 32(3):29:1--29:13, 2013.

\bibitem{DBLP:conf/icml/LehtinenMHLKAA18}
Jaakko Lehtinen, Jacob Munkberg, Jon Hasselgren, Samuli Laine, Tero Karras, Miika Aittala, and Timo Aila.
\newblock Noise2noise: Learning image restoration without clean data.
\newblock In Jennifer~G. Dy and Andreas Krause, editors, {\em International Conference on Machine Learning}, volume~80, pages 2971--2980, 2018.

\bibitem{lin2020sdfsrn}
Chen-Hsuan Lin, Chaoyang Wang, and Simon Lucey.
\newblock {SDF-SRN}: Learning signed distance {3D} object reconstruction from static images.
\newblock In {\em Advances in Neural Information Processing Systems}, 2020.

\bibitem{Lin_2021_CVPR}
Cheng Lin, Changjian Li, Yuan Liu, Nenglun Chen, Yi-King Choi, and Wenping Wang.
\newblock {Point2Skeleton}: Learning skeletal representations from point clouds.
\newblock In {\em Proceedings of the IEEE/CVF Conference on Computer Vision and Pattern Recognition}, pages 4277--4286, 2021.

\bibitem{liu2020meshing}
Minghua Liu, Xiaoshuai Zhang, and Hao Su.
\newblock Meshing point clouds with predicted intrinsic-extrinsic ratio guidance.
\newblock In {\em European Conference on Computer vision}, 2020.

\bibitem{DIST2019SDFRcvpr}
Shaohui Liu, Yinda Zhang, Songyou Peng, Boxin Shi, Marc Pollefeys, and Zhaopeng Cui.
\newblock {DIST}: Rendering deep implicit signed distance function with differentiable sphere tracing.
\newblock In {\em IEEE Conference on Computer Vision and Pattern Recognition}, 2020.

\bibitem{Liu2021MLS}
Shi-Lin Liu, Hao-Xiang Guo, Hao Pan, Pengshuai Wang, Xin Tong, and Yang Liu.
\newblock Deep implicit moving least-squares functions for {3D} reconstruction.
\newblock In {\em IEEE Conference on Computer Vision and Pattern Recognition}, 2021.

\bibitem{shichenNIPS}
Shichen Liu, Shunsuke Saito, Weikai Chen, and Hao Li.
\newblock Learning to infer implicit surfaces without 3{D} supervision.
\newblock In {\em Advances in Neural Information Processing Systems}, 2019.

\bibitem{Lorensen87marchingcubes}
William~E. Lorensen and Harvey~E. Cline.
\newblock Marching cubes: A high resolution {3D} surface construction algorithm.
\newblock {\em Computer Graphics}, 21(4):163--169, 1987.

\bibitem{DBLP:conf/mm/LuoH20}
Shitong Luo and Wei Hu.
\newblock Differentiable manifold reconstruction for point cloud denoising.
\newblock In {\em {ACM} International Conference on Multimedia}, pages 1330--1338. {ACM}, 2020.

\bibitem{luo2021score}
Shitong Luo and Wei Hu.
\newblock Score-based point cloud denoising.
\newblock In {\em Proceedings of the IEEE/CVF International Conference on Computer Vision}, pages 4583--4592, 2021.

\bibitem{Zhizhong2021icml}
Baorui Ma, Zhizhong Han, Yu-Shen Liu, and Matthias Zwicker.
\newblock Neural-pull: Learning signed distance functions from point clouds by learning to pull space onto surfaces.
\newblock In {\em International Conference on Machine Learning}, 2021.

\bibitem{DBLP:conf/cvpr/MaLH22}
Baorui Ma, Yu{-}Shen Liu, and Zhizhong Han.
\newblock Reconstructing surfaces for sparse point clouds with on-surface priors.
\newblock In {\em {IEEE} Conference on Computer Vision and Pattern Recognition}, pages 6305--6315, 2022.

\bibitem{BaoruiNoise2NoiseMapping}
Baorui Ma, Yu-Shen Liu, and Zhizhong Han.
\newblock Learning signed distance functions from noisy {3D} point clouds via noise to noise mapping.
\newblock In {\em International Conference on Machine Learning}, pages 23338--23357. PMLR, 2023.

\bibitem{onsurfacepriors2022}
Baorui Ma, Yu-Shen Liu, Matthias Zwicker, and Zhizhong Han.
\newblock Reconstructing surfaces for sparse point clouds with on-surface priors.
\newblock In {\em {IEEE} Conference on Computer Vision and Pattern Recognition}, 2022.

\bibitem{DBLP:conf/cvpr/MaLZH22}
Baorui Ma, Yu{-}Shen Liu, Matthias Zwicker, and Zhizhong Han.
\newblock Surface reconstruction from point clouds by learning predictive context priors.
\newblock In {\em {IEEE} Conference on Computer Vision and Pattern Recognition}, pages 6316--6327, 2022.

\bibitem{predictivecontextpriors2022}
Baorui Ma, Yu-Shen Liu, Matthias Zwicker, and Zhizhong Han.
\newblock Surface reconstruction from point clouds by learning predictive context priors.
\newblock In {\em {IEEE} Conference on Computer Vision and Pattern Recognition}, 2022.

\bibitem{BaoruiTowards}
Baorui Ma, Junsheng Zhou, Yu-Shen Liu, and Zhizhong Han.
\newblock Towards better gradient consistency for neural signed distance functions via level set alignment.
\newblock In {\em Proceedings of the IEEE/CVF Conference on Computer Vision and Pattern Recognition}, pages 17724--17734, 2023.

\bibitem{Mao_2024_CVPR}
Aihua Mao, Biao Yan, Zijing Ma, and Ying He.
\newblock Denoising point clouds in latent space via graph convolution and invertible neural network.
\newblock In {\em Proceedings of the IEEE/CVF Conference on Computer Vision and Pattern Recognition (CVPR)}, pages 5768--5777, June 2024.

\bibitem{DBLP:journals/corr/abs-2105-02788}
Julien N.~P. Martel, David~B. Lindell, Connor~Z. Lin, Eric~R. Chan, Marco Monteiro, and Gordon Wetzstein.
\newblock {ACORN:} adaptive coordinate networks for neural scene representation.
\newblock {\em CoRR}, abs/2105.02788, 2021.

\bibitem{MeschederNetworks}
Lars Mescheder, Michael Oechsle, Michael Niemeyer, Sebastian Nowozin, and Andreas Geiger.
\newblock Occupancy networks: Learning 3{D} reconstruction in function space.
\newblock In {\em IEEE Conference on Computer Vision and Pattern Recognition}, 2019.

\bibitem{Mi_2020_CVPR}
Zhenxing Mi, Yiming Luo, and Wenbing Tao.
\newblock {SSRNet}: Scalable 3{D} surface reconstruction network.
\newblock In {\em IEEE Conference on Computer Vision and Pattern Recognition}, 2020.

\bibitem{DBLP:journals/corr/abs-1901-06802}
Mateusz Michalkiewicz, Jhony~K. Pontes, Dominic Jack, Mahsa Baktashmotlagh, and Anders~P. Eriksson.
\newblock Deep level sets: Implicit surface representations for 3{D} shape inference.
\newblock {\em CoRR}, abs/1901.06802, 2019.

\bibitem{mildenhall2020nerf}
Ben Mildenhall, Pratul~P. Srinivasan, Matthew Tancik, Jonathan~T. Barron, Ravi Ramamoorthi, and Ren Ng.
\newblock {NeRF}: Representing scenes as neural radiance fields for view synthesis.
\newblock In {\em European Conference on Computer Vision}, 2020.

\bibitem{Volumetric2019SDFRcvpr}
Michael Niemeyer, Lars Mescheder, Michael Oechsle, and Andreas Geiger.
\newblock Differentiable volumetric rendering: Learning implicit 3{D} representations without 3{D} supervision.
\newblock In {\em IEEE Conference on Computer Vision and Pattern Recognition}, 2020.

\bibitem{Oechsle2021ICCV}
Michael Oechsle, Songyou Peng, and Andreas Geiger.
\newblock {UNISURF}: Unifying neural implicit surfaces and radiance fields for multi-view reconstruction.
\newblock In {\em International Conference on Computer Vision}, 2021.

\bibitem{aminie2022}
Amine Ouasfi and Adnane Boukhayma.
\newblock Few 'zero level set'-shot learning of shape signed distance functions in feature space.
\newblock In {\em European Conference on Computer Vision}, 2022.

\bibitem{NEURIPS2023_525c95ff}
Amine Ouasfi and Adnane Boukhayma.
\newblock Robustifying generalizable implicit shape networks with a tunable non-parametric model.
\newblock In {\em Advances in Neural Information Processing Systems}, 2023.

\bibitem{ouasfi2023mixingdenoising}
Amine Ouasfi and Adnane Boukhayma.
\newblock Mixing-denoising generalizable occupancy networks.
\newblock In {\em International Conference on 3D Vision}, 2024.

\bibitem{ouasfi2024unsupervised}
Amine Ouasfi and Adnane Boukhayma.
\newblock Unsupervised occupancy learning from sparse point cloud.
\newblock In {\em {IEEE} Conference on Computer Vision and Pattern Recognition}, 2024.

\bibitem{Park_2019_CVPR}
Jeong~Joon Park, Peter Florence, Julian Straub, Richard Newcombe, and Steven Lovegrove.
\newblock {DeepSDF}: Learning continuous signed distance functions for shape representation.
\newblock In {\em IEEE Conference on Computer Vision and Pattern Recognition}, 2019.

\bibitem{Peng2021SAP}
Songyou Peng, Chiyu~"Max" Jiang, Yiyi Liao, Michael Niemeyer, Marc Pollefeys, and Andreas Geiger.
\newblock Shape as points: A differentiable poisson solver.
\newblock In {\em Advances in Neural Information Processing Systems}, 2021.

\bibitem{DBLP:conf/eccv/PengNMP020}
Songyou Peng, Michael Niemeyer, Lars~M. Mescheder, Marc Pollefeys, and Andreas Geiger.
\newblock Convolutional occupancy networks.
\newblock In {\em European Conference on Computer Vision}, volume 12348, pages 523--540, 2020.

\bibitem{DBLP:conf/eccv/PistilliFVM20}
Francesca Pistilli, Giulia Fracastoro, Diego Valsesia, and Enrico Magli.
\newblock Learning graph-convolutional representations for point cloud denoising.
\newblock In {\em European Conference on Computer Vision}, volume 12365, pages 103--118, 2020.

\bibitem{DBLP:journals/cgf/RakotosaonaBGMO20}
Marie{-}Julie Rakotosaona, Vittorio~La Barbera, Paul Guerrero, Niloy~J. Mitra, and Maks Ovsjanikov.
\newblock Pointcleannet: Learning to denoise and remove outliers from dense point clouds.
\newblock {\em Computer Graphics Forum}, 39(1):185--203, 2020.

\bibitem{rematasICML21}
Konstantinos Rematas, Ricardo Martin-Brualla, and Vittorio Ferrari.
\newblock Sharf: Shape-conditioned radiance fields from a single view.
\newblock In {\em International Conference on Machine Learning}, 2021.

\bibitem{DBLP:conf/icpram/SernaMGD14}
Andr{\'{e}}s Serna, Beatriz Marcotegui, Fran{\c{c}}ois Goulette, and Jean{-}Emmanuel Deschaud.
\newblock Paris-rue-madame database - {A} {3D} mobile laser scanner dataset for benchmarking urban detection, segmentation and classification methods.
\newblock In {\em International Conference on Pattern Recognition Applications and Methods}, pages 819--824, 2014.

\bibitem{shim2024ditto}
Jaehyeok Shim and Kyungdon Joo.
\newblock {DITTO}: Dual and integrated latent topologies for implicit 3d reconstruction.
\newblock {\em arXiv preprint arXiv:2403.05005}, 2024.

\bibitem{sitzmann2019siren}
Vincent Sitzmann, Julien~N.P. Martel, Alexander~W. Bergman, David~B. Lindell, and Gordon Wetzstein.
\newblock Implicit neural representations with periodic activation functions.
\newblock In {\em Advances in Neural Information Processing Systems}, 2020.

\bibitem{sitzmann2019srns}
Vincent Sitzmann, Michael Zollh{\"o}fer, and Gordon Wetzstein.
\newblock Scene representation networks: Continuous 3{D}-structure-aware neural scene representations.
\newblock In {\em Advances in Neural Information Processing Systems}, 2019.

\bibitem{Peng2020ECCV}
Peng Songyou, Niemeyer Michael, Mescheder Lars, Pollefeys Marc, and Geiger Andreas.
\newblock Convolutional occupancy networks.
\newblock In {\em European Conference on Computer Vision}, 2020.

\bibitem{takikawa2021nglod}
Towaki Takikawa, Joey Litalien, Kangxue Yin, Karsten Kreis, Charles Loop, Derek Nowrouzezahrai, Alec Jacobson, Morgan McGuire, and Sanja Fidler.
\newblock Neural geometric level of detail: Real-time rendering with implicit {3D} shapes.
\newblock In {\em IEEE Conference on Computer Vision and Pattern Recognition}, 2021.

\bibitem{tang2021sign}
Jiapeng Tang, Jiabao Lei, Dan Xu, Feiying Ma, Kui Jia, and Lei Zhang.
\newblock {SA-ConvONet}: Sign-agnostic optimization of convolutional occupancy networks.
\newblock In {\em Proceedings of the IEEE/CVF International Conference on Computer Vision}, 2021.

\bibitem{Tatarchenko_2019_CVPR}
Maxim Tatarchenko, Stephan~R. Richter, Rene Ranftl, Zhuwen Li, Vladlen Koltun, and Thomas Brox.
\newblock What do single-view {3D} reconstruction networks learn?
\newblock In {\em The IEEE Conference on Computer Vision and Pattern Recognition}, 2019.

\bibitem{Tretschk2020PatchNets}
Edgar Tretschk, Ayush Tewari, Vladislav Golyanik, Michael Zollh\"{o}fer, Carsten Stoll, and Christian Theobalt.
\newblock {PatchNets: Patch-Based Generalizable Deep Implicit 3D Shape Representations}.
\newblock {\em European Conference on Computer Vision}, 2020.

\bibitem{Vicini2022sdf}
Delio Vicini, Sébastien Speierer, and Wenzel Jakob.
\newblock Differentiable signed distance function rendering.
\newblock {\em ACM Transactions on Graphics}, 41(4):125:1--125:18, 2022.

\bibitem{wang2022neuris}
Jiepeng Wang, Peng Wang, Xiaoxiao Long, Christian Theobalt, Taku Komura, Lingjie Liu, and Wenping Wang.
\newblock {NeuRIS}: Neural reconstruction of indoor scenes using normal priors.
\newblock In {\em European Conference on Computer Vision}, 2022.

\bibitem{neuslingjie}
Peng Wang, Lingjie Liu, Yuan Liu, Christian Theobalt, Taku Komura, and Wenping Wang.
\newblock {NeuS}: Learning neural implicit surfaces by volume rendering for multi-view reconstruction.
\newblock In {\em Advances in Neural Information Processing Systems}, pages 27171--27183, 2021.

\bibitem{wang2020deep}
Peng-Shuai Wang, Yang Liu, and Xin Tong.
\newblock Deep octree-based cnns with output-guided skip connections for 3d shape and scene completion.
\newblock In {\em Proceedings of the IEEE/CVF Conference on Computer Vision and Pattern Recognition Workshops}, pages 266--267, 2020.

\bibitem{NEURIPS2023_c87bd584}
Ruian Wang, Zixiong Wang, Yunxiao Zhang, Shuangmin Chen, Shiqing Xin, Changhe Tu, and Wenping Wang.
\newblock Aligning gradient and hessian for neural signed distance function.
\newblock In {\em Advances in Neural Information Processing Systems}, volume~36, pages 63515--63528, 2023.

\bibitem{yiqunhfSDF}
Yiqun Wang, Ivan Skorokhodov, and Peter Wonka.
\newblock {HF-NeuS}: Improved surface reconstruction using high-frequency details.
\newblock 2022.

\bibitem{Wang_2023_CVPR}
Zhen Wang, Shijie Zhou, Jeong~Joon Park, Despoina Paschalidou, Suya You, Gordon Wetzstein, Leonidas Guibas, and Achuta Kadambi.
\newblock Alto: Alternating latent topologies for implicit 3d reconstruction.
\newblock In {\em Proceedings of the IEEE/CVF Conference on Computer Vision and Pattern Recognition (CVPR)}, pages 259--270, 2023.

\bibitem{Williams_2022_CVPR}
Francis Williams, Zan Gojcic, Sameh Khamis, Denis Zorin, Joan Bruna, Sanja Fidler, and Or~Litany.
\newblock Neural fields as learnable kernels for {3D} reconstruction.
\newblock In {\em Proceedings of the IEEE/CVF Conference on Computer Vision and Pattern Recognition}, pages 18500--18510, 2022.

\bibitem{Williams_2019_CVPR}
Francis Williams, Teseo Schneider, Claudio Silva, Denis Zorin, Joan Bruna, and Daniele Panozzo.
\newblock Deep geometric prior for surface reconstruction.
\newblock In {\em IEEE Conference on Computer Vision and Pattern Recognition}, 2019.

\bibitem{DBLP:conf/cvpr/WilliamsTBZ21}
Francis Williams, Matthew Trager, Joan Bruna, and Denis Zorin.
\newblock Neural splines: Fitting 3{D} surfaces with infinitely-wide neural networks.
\newblock In {\em {IEEE} Conference on Computer Vision and Pattern Recognition}, pages 9949--9958, 2021.

\bibitem{DBLP:journals/cgf/WuS20}
Yunjie Wu and Zhengxing Sun.
\newblock {DFR:} differentiable function rendering for learning {3D} generation from images.
\newblock {\em Computer Graphics Forum}, 39(5):241--252, 2020.

\bibitem{zhizhongiccv2021completing}
Peng Xiang, Xin Wen, Yu-Shen Liu, Yan-Pei Cao, Pengfei Wan, Wen Zheng, and Zhizhong Han.
\newblock {SnowflakeNet}: Point cloud completion by snowflake point deconvolution with skip-transformer.
\newblock In {\em {IEEE} International Conference on Computer Vision}, 2021.

\bibitem{NEURIPS2023_2d6336c1}
Huizong Yang, Yuxin Sun, Ganesh Sundaramoorthi, and Anthony Yezzi.
\newblock {StEik}: Stabilizing the optimization of neural signed distance functions and finer shape representation.
\newblock In {\em Advances in Neural Information Processing Systems}, 2023.

\bibitem{yariv2021volume}
Lior Yariv, Jiatao Gu, Yoni Kasten, and Yaron Lipman.
\newblock Volume rendering of neural implicit surfaces.
\newblock In {\em Advances in Neural Information Processing Systems}, 2021.

\bibitem{yariv2020multiview}
Lior Yariv, Yoni Kasten, Dror Moran, Meirav Galun, Matan Atzmon, Basri Ronen, and Yaron Lipman.
\newblock Multiview neural surface reconstruction by disentangling geometry and appearance.
\newblock {\em Advances in Neural Information Processing Systems}, 33, 2020.

\bibitem{yifan2020isopoints}
Wang Yifan, Shihao Wu, Cengiz Oztireli, and Olga Sorkine-Hornung.
\newblock {Iso-Points}: Optimizing neural implicit surfaces with hybrid representations.
\newblock {\em CoRR}, abs/2012.06434, 2020.

\bibitem{Yu2022MonoSDF}
Zehao Yu, Songyou Peng, Michael Niemeyer, Torsten Sattler, and Andreas Geiger.
\newblock {MonoSDF}: Exploring monocular geometric cues for neural implicit surface reconstruction.
\newblock 2022.

\bibitem{prior2019SDFRcvpr}
Sergey Zakharov, Wadim Kehl, Arjun Bhargava, and Adrien Gaidon.
\newblock Autolabeling 3{D} objects with differentiable rendering of sdf shape priors.
\newblock In {\em IEEE Conference on Computer Vision and Pattern Recognition}, 2020.

\bibitem{zhao2020signagnostic}
Wenbin Zhao, Jiabao Lei, Yuxin Wen, Jianguo Zhang, and Kui Jia.
\newblock Sign-agnostic implicit learning of surface self-similarities for shape modeling and reconstruction from raw point clouds.
\newblock {\em CoRR}, abs/2012.07498, 2020.

\bibitem{Zhou_2023_ICCV}
Junsheng Zhou, Baorui Ma, Shujuan Li, Yu-Shen Liu, and Zhizhong Han.
\newblock Learning a more continuous zero level set in unsigned distance fields through level set projection.
\newblock In {\em Proceedings of the IEEE/CVF International Conference on Computer Vision}, pages 3181--3192, 2023.

\bibitem{Zhou2022CAP-UDF}
Junsheng Zhou, Baorui Ma, Yu-Shen Liu, Yi~Fang, and Zhizhong Han.
\newblock Learning consistency-aware unsigned distance functions progressively from raw point clouds.
\newblock In {\em Advances in Neural Information Processing Systems}, 2022.

\bibitem{DBLP:journals/tog/ZhouK13}
Qian{-}Yi Zhou and Vladlen Koltun.
\newblock Dense scene reconstruction with points of interest.
\newblock {\em {ACM} Transactions on Graphics}, 32(4):112:1--112:8, 2013.

\bibitem{Zhu_2024_WACV}
Runsong Zhu, Di~Kang, Ka-Hei Hui, Yue Qian, Shi Qiu, Zhen Dong, Linchao Bao, Pheng-Ann Heng, and Chi-Wing Fu.
\newblock Ssp: Semi-signed prioritized neural fitting for surface reconstruction from unoriented point clouds.
\newblock In {\em Proceedings of the IEEE/CVF Winter Conference on Applications of Computer Vision (WACV)}, pages 3769--3778, 2024.

\end{thebibliography}
}


\appendix

\section{Appendix}

\subsection{Limitations}
Our method is still limited to too large noises. For noises that corrupted shapes too much, our method still produces bad results. One direction for our future work is to improve our prior, so that we could have a better sense of a shape even under large noises.

\subsection{Detailed Comparisons on 3D Scene}
We detail our evaluations on each scene in 3D scene dataset in Tab.~\ref{table:3DScene1}. The comparisons highlight our advantages in each scene.

\begin{table}[h]
\vspace{-0.1in}
\centering
\resizebox{\linewidth}{!}{
    \begin{tabular}{c|c|c|c|c|c|c|c}  
     \hline
     Name&Metrics&COcc~\cite{DBLP:conf/eccv/PengNMP020}&LIG~\cite{jiang2020lig}&DeepLS\cite{DBLP:conf/eccv/ChabraLISSLN20}&NP~\cite{Zhizhong2021icml}&N2NM~\cite{BaoruiNoise2NoiseMapping}&Ours\\
     \hline
     \multirow{3}{*}{Burghers}&$CD_{L2}\times 1000$&27.46&3.055&\textbf{0.401}&1.204&0.504&0.429\\
     &$CD_{L1}$&0.079&0.045&0.017&0.031&0.020&\textbf{0.016}\\
     &NC&0.907&0.835&0.920&0.905&0.925&\textbf{0.939}\\
     \hline
     \multirow{3}{*}{Lounge}&$CD_{L2}\times 1000$&9.540&9.672&6.103&1.079&0.602&\textbf{0.333}\\
     &$CD_{L1}$&0.046&0.056&0.053&0.019&0.016&\textbf{0.014}\\
     &NC&0.894&0.833&0.848&0.910&0.923&\textbf{0.935}\\
     \hline
     \multirow{3}{*}{Copyroom}&$CD_{L2}\times 1000$&10.97&3.610&0.609&5.795&0.442&\textbf{0.389}\\
     &$CD_{L1}$&0.045&0.036&0.021&0.036&\textbf{0.016}&\textbf{0.016}\\
     &NC&0.892&0.810&0.901&0.862&0.903&\textbf{0.916}\\
     \hline
     \multirow{3}{*}{Stonewall}&$CD_{L2}\times 1000$&20.46&5.032&0.320&0.983&0.330&\textbf{0.313}\\
     &$CD_{L1}$&0.069&0.042&0.015&0.029&0.020&\textbf{0.015}\\
     &NC&0.905&0.879&0.954&0.930&0.951&\textbf{0.961}\\
     \hline
     \multirow{3}{*}{Totepole}&$CD_{L2}\times 1000$&2.054&9.580&0.601&1.513&0.657&\textbf{0.482}\\
     &$CD_{L1}$&0.021&0.062&\textbf{0.017}&0.054&0.023&0.020\\
     &NC&0.943&0.887&0.950&0.893&0.945&\textbf{0.957}\\
     \hline
   \end{tabular}}
   \vspace{-0.0in}
   \caption{Numerical Comparisons on 3D Scene dataset in terms of $CD_{L1}$, $CD_{L2}$ and NC.}
   \label{table:3DScene1}
   \vspace{-0.1in}
\end{table}

\subsection{More Results}
\begin{figure}[h]
\vspace{-0.0in}
  \centering
   \includegraphics[width=\linewidth]{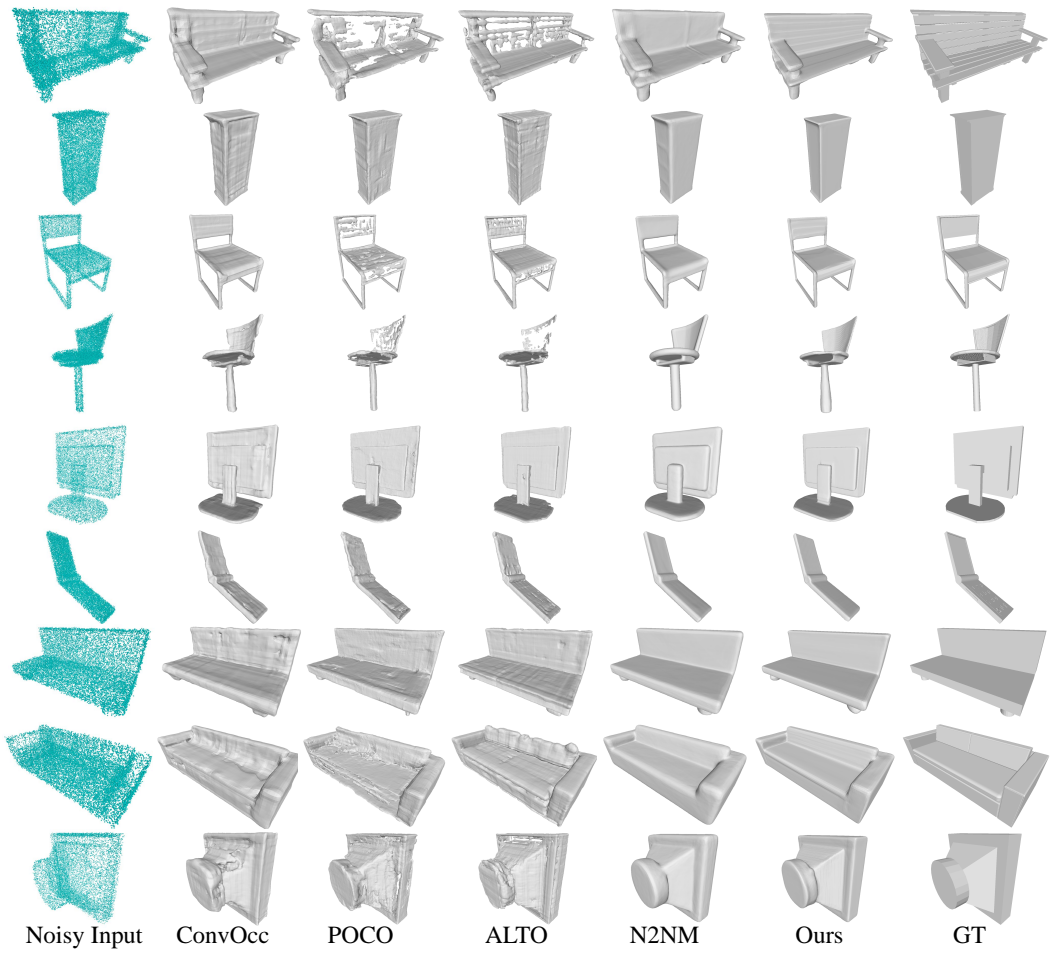}
\vspace{-0.0in}
\caption{\label{fig:shapenet_app}Comparison in surface reconstruction on ShapeNet.}
\vspace{-0.0in}
\end{figure}

\begin{figure}[h]
\vspace{-0.0in}
  \centering
   \includegraphics[width=\linewidth]{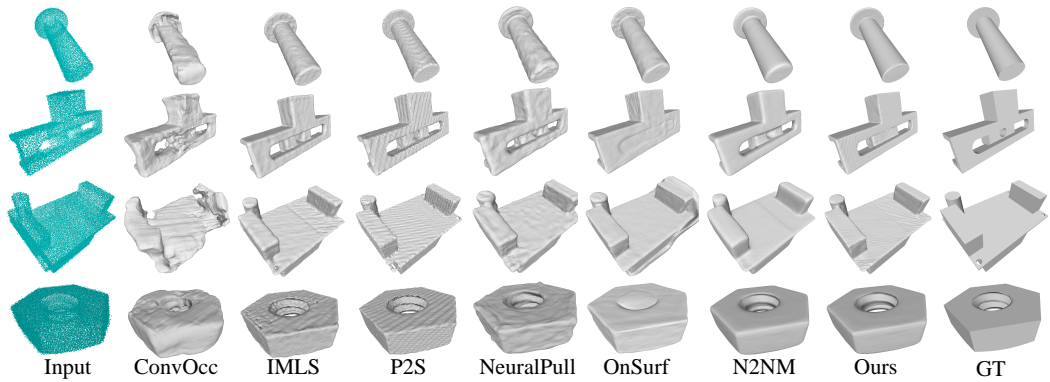}
\vspace{-0.0in}
\caption{\label{fig:ABC_app}Comparison in surface reconstruction on ABC.}
\vspace{-0.0in}
\end{figure}

\begin{figure}[h]
  \centering
   \includegraphics[width=\linewidth]{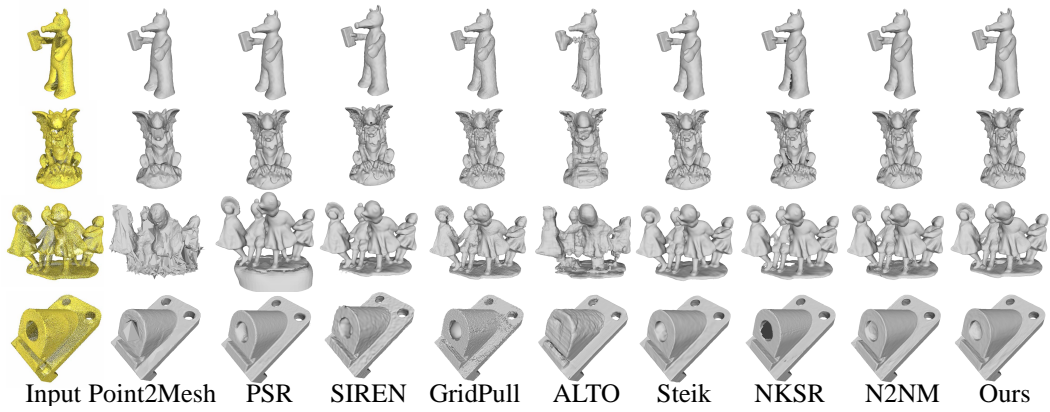}
\caption{\label{fig:srb_app}Comparison in surface reconstruction on SRB.}
\end{figure}

\begin{figure}[h]
  \centering
   \includegraphics[width=\linewidth]{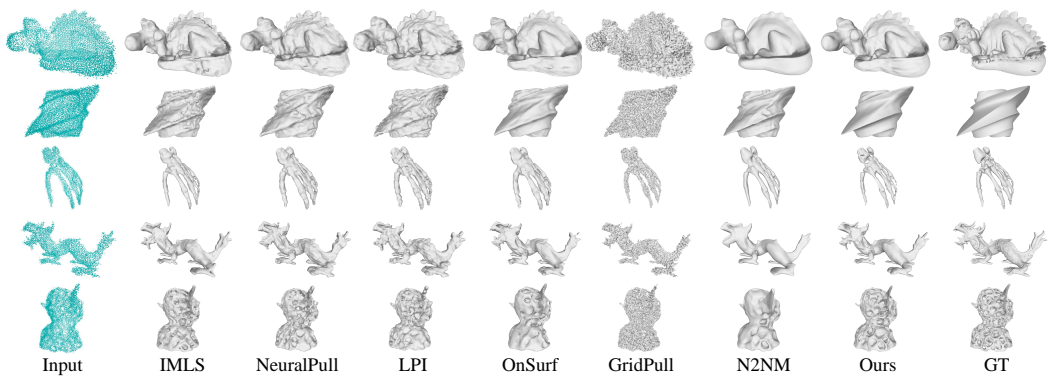}
\caption{\label{fig:famous_app1}Comparison in surface reconstruction on FAMOUS.}
\end{figure}

\begin{figure}[h]
\vspace{-0.0in}
  \centering
   \includegraphics[width=\linewidth]{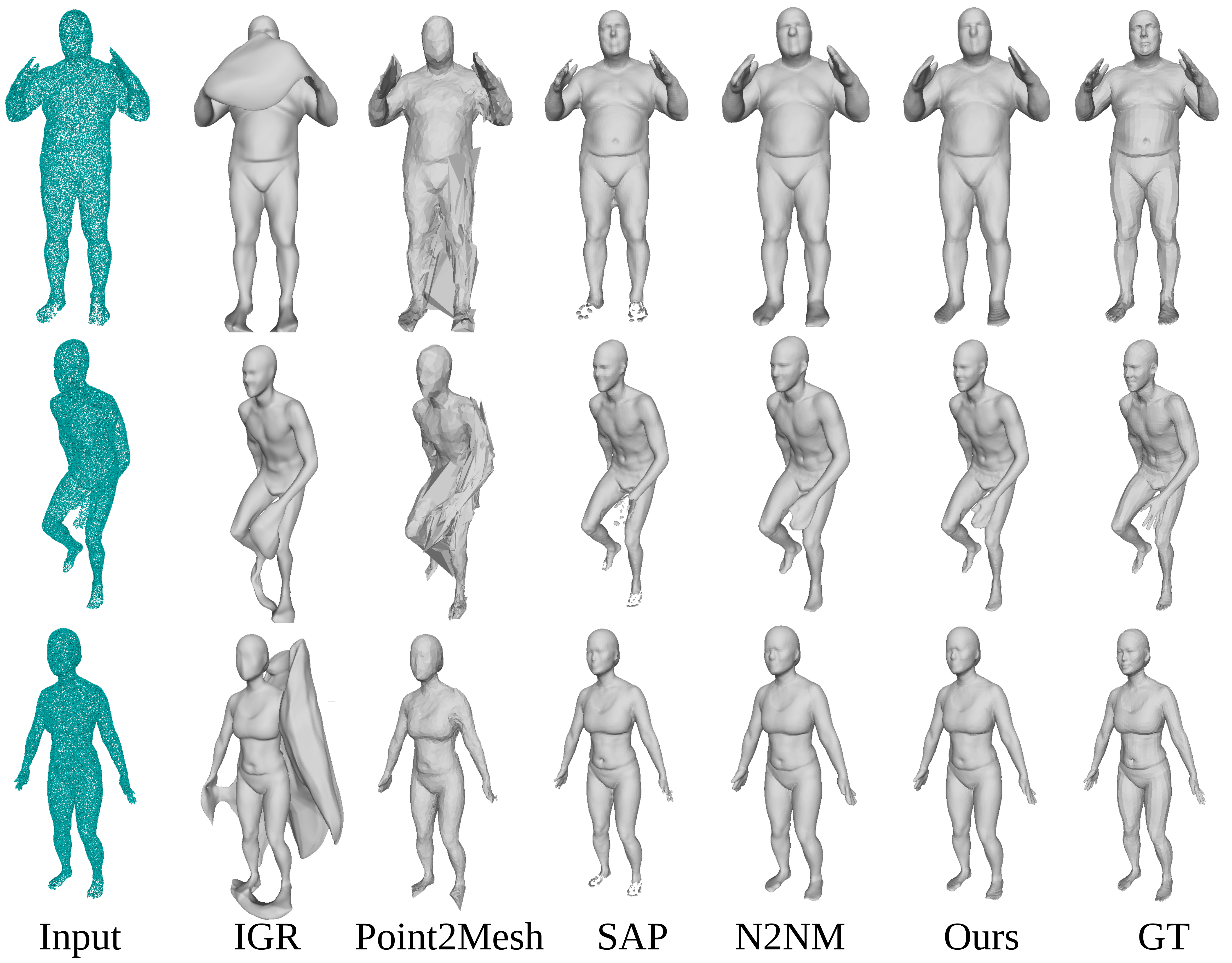}
\caption{\label{fig:dfaust_app}Comparison in surface reconstruction on D-FAUST.}
\end{figure}

\begin{figure}[h]
  \centering
   \includegraphics[width=0.8\linewidth]{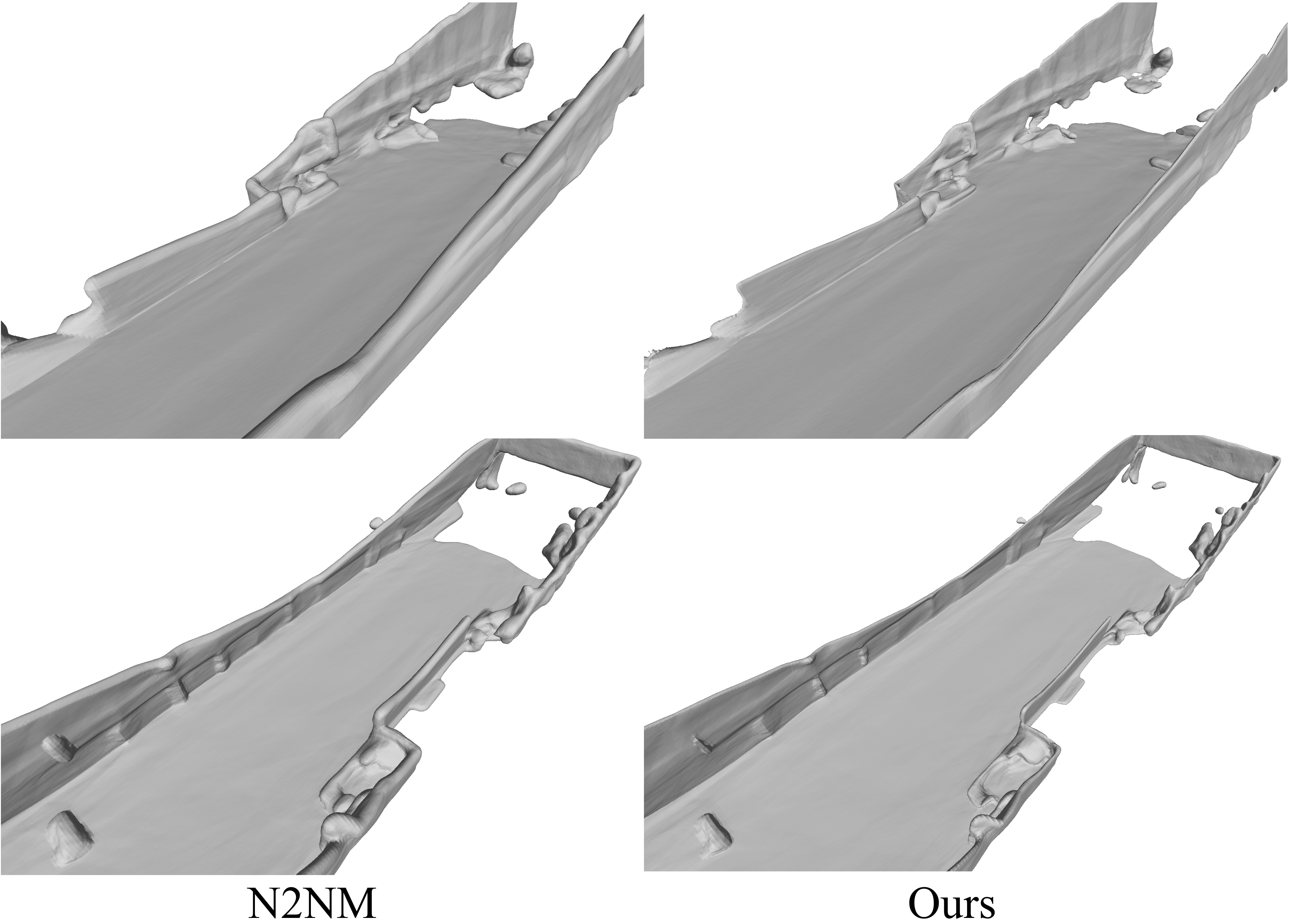}
\caption{\label{fig:nuscene_app}Comparison in surface reconstruction on nuScenes.
}
\end{figure}

We visualize more surface reconstruction results under ShapeNet~\cite{shapenet2015}, ABC~\cite{ErlerEtAl:Points2Surf:ECCV:2020}, Surface Reconstruction Benchmark (SRB)~\cite{Williams_2019_CVPR}, FAMOUS~\cite{ErlerEtAl:Points2Surf:ECCV:2020}, D-FAUST~\cite{dfaust:CVPR:2017} and nuScenes~\cite{nuscenes} in Fig.~\ref{fig:shapenet_app}, Fig.~\ref{fig:ABC_app}, Fig.~\ref{fig:srb_app}, Fig.~\ref{fig:famous_app1}, Fig.~\ref{fig:dfaust_app} and Fig.~\ref{fig:nuscene_app}. 

\clearpage
\newpage

\section*{NeurIPS Paper Checklist}

\begin{enumerate}

\item {\bf Claims}
    \item[] Question: Do the main claims made in the abstract and introduction accurately reflect the paper's contributions and scope?
    \item[] Answer: \answerYes{} 
    \item[] Justification: Our main claims made in the abstract and introduction accurately reflect the paper’s contributions and scope. 
    \item[] Guidelines:
    \begin{itemize}
        \item The answer NA means that the abstract and introduction do not include the claims made in the paper.
        \item The abstract and/or introduction should clearly state the claims made, including the contributions made in the paper and important assumptions and limitations. A No or NA answer to this question will not be perceived well by the reviewers. 
        \item The claims made should match theoretical and experimental results, and reflect how much the results can be expected to generalize to other settings. 
        \item It is fine to include aspirational goals as motivation as long as it is clear that these goals are not attained by the paper. 
    \end{itemize}

\item {\bf Limitations}
    \item[] Question: Does the paper discuss the limitations of the work performed by the authors?
    \item[] Answer: \answerYes{} 
    \item[] Justification: We discuss the limitations in the Appendix A.1.
    \item[] Guidelines:
    \begin{itemize}
        \item The answer NA means that the paper has no limitation while the answer No means that the paper has limitations, but those are not discussed in the paper. 
        \item The authors are encouraged to create a separate "Limitations" section in their paper.
        \item The paper should point out any strong assumptions and how robust the results are to violations of these assumptions (e.g., independence assumptions, noiseless settings, model well-specification, asymptotic approximations only holding locally). The authors should reflect on how these assumptions might be violated in practice and what the implications would be.
        \item The authors should reflect on the scope of the claims made, e.g., if the approach was only tested on a few datasets or with a few runs. In general, empirical results often depend on implicit assumptions, which should be articulated.
        \item The authors should reflect on the factors that influence the performance of the approach. For example, a facial recognition algorithm may perform poorly when image resolution is low or images are taken in low lighting. Or a speech-to-text system might not be used reliably to provide closed captions for online lectures because it fails to handle technical jargon.
        \item The authors should discuss the computational efficiency of the proposed algorithms and how they scale with dataset size.
        \item If applicable, the authors should discuss possible limitations of their approach to address problems of privacy and fairness.
        \item While the authors might fear that complete honesty about limitations might be used by reviewers as grounds for rejection, a worse outcome might be that reviewers discover limitations that aren't acknowledged in the paper. The authors should use their best judgment and recognize that individual actions in favor of transparency play an important role in developing norms that preserve the integrity of the community. Reviewers will be specifically instructed to not penalize honesty concerning limitations.
    \end{itemize}

\item {\bf Theory Assumptions and Proofs}
    \item[] Question: For each theoretical result, does the paper provide the full set of assumptions and a complete (and correct) proof?
    \item[] Answer: \answerNA{} 
    \item[] Justification: We describe in Method one of the core contributions of the local noise-to-noise mapping, and although there is no theory or theorem in it, we verify its validity and reasonableness in our experiments. 
    \item[] Guidelines:
    \begin{itemize}
        \item The answer NA means that the paper does not include theoretical results. 
        \item All the theorems, formulas, and proofs in the paper should be numbered and cross-referenced.
        \item All assumptions should be clearly stated or referenced in the statement of any theorems.
        \item The proofs can either appear in the main paper or the supplemental material, but if they appear in the supplemental material, the authors are encouraged to provide a short proof sketch to provide intuition. 
        \item Inversely, any informal proof provided in the core of the paper should be complemented by formal proofs provided in appendix or supplemental material.
        \item Theorems and Lemmas that the proof relies upon should be properly referenced. 
    \end{itemize}

    \item {\bf Experimental Result Reproducibility}
    \item[] Question: Does the paper fully disclose all the information needed to reproduce the main experimental results of the paper to the extent that it affects the main claims and/or conclusions of the paper (regardless of whether the code and data are provided or not)?
    \item[] Answer: \answerYes{} 
    \item[] Justification: We provide detailed information in reproducing our methods in Implementation Details of Section 3. 
    \item[] Guidelines:
    \begin{itemize}
        \item The answer NA means that the paper does not include experiments.
        \item If the paper includes experiments, a No answer to this question will not be perceived well by the reviewers: Making the paper reproducible is important, regardless of whether the code and data are provided or not.
        \item If the contribution is a dataset and/or model, the authors should describe the steps taken to make their results reproducible or verifiable. 
        \item Depending on the contribution, reproducibility can be accomplished in various ways. For example, if the contribution is a novel architecture, describing the architecture fully might suffice, or if the contribution is a specific model and empirical evaluation, it may be necessary to either make it possible for others to replicate the model with the same dataset, or provide access to the model. In general. releasing code and data is often one good way to accomplish this, but reproducibility can also be provided via detailed instructions for how to replicate the results, access to a hosted model (e.g., in the case of a large language model), releasing of a model checkpoint, or other means that are appropriate to the research performed.
        \item While NeurIPS does not require releasing code, the conference does require all submissions to provide some reasonable avenue for reproducibility, which may depend on the nature of the contribution. For example
        \begin{enumerate}
            \item If the contribution is primarily a new algorithm, the paper should make it clear how to reproduce that algorithm.
            \item If the contribution is primarily a new model architecture, the paper should describe the architecture clearly and fully.
            \item If the contribution is a new model (e.g., a large language model), then there should either be a way to access this model for reproducing the results or a way to reproduce the model (e.g., with an open-source dataset or instructions for how to construct the dataset).
            \item We recognize that reproducibility may be tricky in some cases, in which case authors are welcome to describe the particular way they provide for reproducibility. In the case of closed-source models, it may be that access to the model is limited in some way (e.g., to registered users), but it should be possible for other researchers to have some path to reproducing or verifying the results.
        \end{enumerate}
    \end{itemize}

\item {\bf Open access to data and code}
    \item[] Question: Does the paper provide open access to the data and code, with sufficient instructions to faithfully reproduce the main experimental results, as described in supplemental material?
    \item[] Answer: \answerYes{} 
    \item[] Justification: We provide our demonstration code as a part of our supplementary materials. We will release our source code, data and sufficient instructions upon acceptance. 
    \item[] Guidelines:
    \begin{itemize}
        \item The answer NA means that paper does not include experiments requiring code.
        \item Please see the NeurIPS code and data submission guidelines (\url{https://nips.cc/public/guides/CodeSubmissionPolicy}) for more details.
        \item While we encourage the release of code and data, we understand that this might not be possible, so “No” is an acceptable answer. Papers cannot be rejected simply for not including code, unless this is central to the contribution (e.g., for a new open-source benchmark).
        \item The instructions should contain the exact command and environment needed to run to reproduce the results. See the NeurIPS code and data submission guidelines (\url{https://nips.cc/public/guides/CodeSubmissionPolicy}) for more details.
        \item The authors should provide instructions on data access and preparation, including how to access the raw data, preprocessed data, intermediate data, and generated data, etc.
        \item The authors should provide scripts to reproduce all experimental results for the new proposed method and baselines. If only a subset of experiments are reproducible, they should state which ones are omitted from the script and why.
        \item At submission time, to preserve anonymity, the authors should release anonymized versions (if applicable).
        \item Providing as much information as possible in supplemental material (appended to the paper) is recommended, but including URLs to data and code is permitted.
    \end{itemize}

\item {\bf Experimental Setting/Details}
    \item[] Question: Does the paper specify all the training and test details (e.g., data splits, hyperparameters, how they were chosen, type of optimizer, etc.) necessary to understand the results?
    \item[] Answer: \answerYes{} 
    \item[] Justification: We provide all the training and test details for shapes and scenes in Section 4. 
    \item[] Guidelines:
    \begin{itemize}
        \item The answer NA means that the paper does not include experiments.
        \item The experimental setting should be presented in the core of the paper to a level of detail that is necessary to appreciate the results and make sense of them.
        \item The full details can be provided either with the code, in appendix, or as supplemental material.
    \end{itemize}

\item {\bf Experiment Statistical Significance}
    \item[] Question: Does the paper report error bars suitably and correctly defined or other appropriate information about the statistical significance of the experiments?
    \item[] Answer: \answerNo{} 
    \item[] Justification:  We report the average performance in terms of several metrics as the experimental results. 
    \item[] Guidelines:
    \begin{itemize}
        \item The answer NA means that the paper does not include experiments.
        \item The authors should answer "Yes" if the results are accompanied by error bars, confidence intervals, or statistical significance tests, at least for the experiments that support the main claims of the paper.
        \item The factors of variability that the error bars are capturing should be clearly stated (for example, train/test split, initialization, random drawing of some parameter, or overall run with given experimental conditions).
        \item The method for calculating the error bars should be explained (closed form formula, call to a library function, bootstrap, etc.)
        \item The assumptions made should be given (e.g., Normally distributed errors).
        \item It should be clear whether the error bar is the standard deviation or the standard error of the mean.
        \item It is OK to report 1-sigma error bars, but one should state it. The authors should preferably report a 2-sigma error bar than state that they have a 96\% CI, if the hypothesis of Normality of errors is not verified.
        \item For asymmetric distributions, the authors should be careful not to show in tables or figures symmetric error bars that would yield results that are out of range (e.g. negative error rates).
        \item If error bars are reported in tables or plots, The authors should explain in the text how they were calculated and reference the corresponding figures or tables in the text.
    \end{itemize}

\item {\bf Experiments Compute Resources}
    \item[] Question: For each experiment, does the paper provide sufficient information on the computer resources (type of compute workers, memory, time of execution) needed to reproduce the experiments?
    \item[] Answer: \answerYes{} 
    \item[] Justification: We report our inference time with other methods in the experiments. 
    \item[] Guidelines:
    \begin{itemize}
        \item The answer NA means that the paper does not include experiments.
        \item The paper should indicate the type of compute workers CPU or GPU, internal cluster, or cloud provider, including relevant memory and storage.
        \item The paper should provide the amount of compute required for each of the individual experimental runs as well as estimate the total compute. 
        \item The paper should disclose whether the full research project required more compute than the experiments reported in the paper (e.g., preliminary or failed experiments that didn't make it into the paper). 
    \end{itemize}
    
\item {\bf Code Of Ethics}
    \item[] Question: Does the research conducted in the paper conform, in every respect, with the NeurIPS Code of Ethics \url{https://neurips.cc/public/EthicsGuidelines}?
    \item[] Answer: \answerYes{} 
    \item[] Justification: The research conducted in this paper conforms in all respects to the NeurIPS Code of Ethics.
    \item[] Guidelines:
    \begin{itemize}
        \item The answer NA means that the authors have not reviewed the NeurIPS Code of Ethics.
        \item If the authors answer No, they should explain the special circumstances that require a deviation from the Code of Ethics.
        \item The authors should make sure to preserve anonymity (e.g., if there is a special consideration due to laws or regulations in their jurisdiction).
    \end{itemize}

\item {\bf Broader Impacts}
    \item[] Question: Does the paper discuss both potential positive societal impacts and negative societal impacts of the work performed?
    \item[] Answer: \answerYes{} 
    \item[] Justification: We discuss the application and potential positive impact of our method in the introduction. 
    \item[] Guidelines:
    \begin{itemize}
        \item The answer NA means that there is no societal impact of the work performed.
        \item If the authors answer NA or No, they should explain why their work has no societal impact or why the paper does not address societal impact.
        \item Examples of negative societal impacts include potential malicious or unintended uses (e.g., disinformation, generating fake profiles, surveillance), fairness considerations (e.g., deployment of technologies that could make decisions that unfairly impact specific groups), privacy considerations, and security considerations.
        \item The conference expects that many papers will be foundational research and not tied to particular applications, let alone deployments. However, if there is a direct path to any negative applications, the authors should point it out. For example, it is legitimate to point out that an improvement in the quality of generative models could be used to generate deepfakes for disinformation. On the other hand, it is not needed to point out that a generic algorithm for optimizing neural networks could enable people to train models that generate Deepfakes faster.
        \item The authors should consider possible harms that could arise when the technology is being used as intended and functioning correctly, harms that could arise when the technology is being used as intended but gives incorrect results, and harms following from (intentional or unintentional) misuse of the technology.
        \item If there are negative societal impacts, the authors could also discuss possible mitigation strategies (e.g., gated release of models, providing defenses in addition to attacks, mechanisms for monitoring misuse, mechanisms to monitor how a system learns from feedback over time, improving the efficiency and accessibility of ML).
    \end{itemize}
    
\item {\bf Safeguards}
    \item[] Question: Does the paper describe safeguards that have been put in place for responsible release of data or models that have a high risk for misuse (e.g., pretrained language models, image generators, or scraped datasets)?
    \item[] Answer: \answerNA{} 
    \item[] Justification: There is no such risk to the paper.
    \item[] Guidelines:
    \begin{itemize}
        \item The answer NA means that the paper poses no such risks.
        \item Released models that have a high risk for misuse or dual-use should be released with necessary safeguards to allow for controlled use of the model, for example by requiring that users adhere to usage guidelines or restrictions to access the model or implementing safety filters. 
        \item Datasets that have been scraped from the Internet could pose safety risks. The authors should describe how they avoided releasing unsafe images.
        \item We recognize that providing effective safeguards is challenging, and many papers do not require this, but we encourage authors to take this into account and make a best faith effort.
    \end{itemize}

\item {\bf Licenses for existing assets}
    \item[] Question: Are the creators or original owners of assets (e.g., code, data, models), used in the paper, properly credited and are the license and terms of use explicitly mentioned and properly respected?
    \item[] Answer: \answerYes{} 
    \item[] Justification: We use open-source datasets and code under their licence. 
    \item[] Guidelines:
    \begin{itemize}
        \item The answer NA means that the paper does not use existing assets.
        \item The authors should cite the original paper that produced the code package or dataset.
        \item The authors should state which version of the asset is used and, if possible, include a URL.
        \item The name of the license (e.g., CC-BY 4.0) should be included for each asset.
        \item For scraped data from a particular source (e.g., website), the copyright and terms of service of that source should be provided.
        \item If assets are released, the license, copyright information, and terms of use in the package should be provided. For popular datasets, \url{paperswithcode.com/datasets} has curated licenses for some datasets. Their licensing guide can help determine the license of a dataset.
        \item For existing datasets that are re-packaged, both the original license and the license of the derived asset (if it has changed) should be provided.
        \item If this information is not available online, the authors are encouraged to reach out to the asset's creators.
    \end{itemize}

\item {\bf New Assets}
    \item[] Question: Are new assets introduced in the paper well documented and is the documentation provided alongside the assets?
    \item[] Answer: \answerNA{} 
    \item[] Justification: The paper does not release new assets. 
    \item[] Guidelines:
    \begin{itemize}
        \item The answer NA means that the paper does not release new assets.
        \item Researchers should communicate the details of the dataset/code/model as part of their submissions via structured templates. This includes details about training, license, limitations, etc. 
        \item The paper should discuss whether and how consent was obtained from people whose asset is used.
        \item At submission time, remember to anonymize your assets (if applicable). You can either create an anonymized URL or include an anonymized zip file.
    \end{itemize}

\item {\bf Crowdsourcing and Research with Human Subjects}
    \item[] Question: For crowdsourcing experiments and research with human subjects, does the paper include the full text of instructions given to participants and screenshots, if applicable, as well as details about compensation (if any)? 
    \item[] Answer: \answerNA{} 
    \item[] Justification: The paper does not involve crowdsourcing nor research with human subjects. 
    \item[] Guidelines:
    \begin{itemize}
        \item The answer NA means that the paper does not involve crowdsourcing nor research with human subjects.
        \item Including this information in the supplemental material is fine, but if the main contribution of the paper involves human subjects, then as much detail as possible should be included in the main paper. 
        \item According to the NeurIPS Code of Ethics, workers involved in data collection, curation, or other labor should be paid at least the minimum wage in the country of the data collector. 
    \end{itemize}

\item {\bf Institutional Review Board (IRB) Approvals or Equivalent for Research with Human Subjects}
    \item[] Question: Does the paper describe potential risks incurred by study participants, whether such risks were disclosed to the subjects, and whether Institutional Review Board (IRB) approvals (or an equivalent approval/review based on the requirements of your country or institution) were obtained?
    \item[] Answer: \answerNA{} 
    \item[] Justification: The paper does not involve crowdsourcing nor research with human subjects. 
    \item[] Guidelines:
    \begin{itemize}
        \item The answer NA means that the paper does not involve crowdsourcing nor research with human subjects.
        \item Depending on the country in which research is conducted, IRB approval (or equivalent) may be required for any human subjects research. If you obtained IRB approval, you should clearly state this in the paper. 
        \item We recognize that the procedures for this may vary significantly between institutions and locations, and we expect authors to adhere to the NeurIPS Code of Ethics and the guidelines for their institution. 
        \item For initial submissions, do not include any information that would break anonymity (if applicable), such as the institution conducting the review.
    \end{itemize}

\end{enumerate}

\end{document}